# An Uncertainty-Informed Framework for Trustworthy Fault Diagnosis in Safety-Critical Applications

Taotao Zhou[1*], Enrique Lopez Droguett[2,3], Ali Mosleh[3], Felix T.S. Chan[4,5]

[1]*Center for Risk and Reliability, University of Maryland, College Park, USA*
[2]*Department of Civil and Environmental Engineering, University of California, Los Angeles, USA*
[3]*Garrick Institute for the Risk Sciences, University of California, Los Angeles, USA*
[4]*Department of Industrial and Systems Engineering, The Hong Kong Polytechnic University, Hung Hom, Kowloon, Hong Kong*
[5]*Centre for Advances in Reliability and Safety Limited (CAiRS), Hong Kong Science Park, New Territories, Hong Kong*

***ABSTRACT***
There has been a growing interest in deep learning-based prognostic and health management (PHM) for building end-to-end maintenance decision support systems, especially due to the rapid development of autonomous systems. However, the low trustworthiness of PHM hinders its applications in safety-critical assets when handling data from an unknown distribution that differs from the training dataset, referred to as the out-of-distribution (OOD) dataset. To bridge this gap, we propose an uncertainty-informed framework to diagnose faults and meanwhile detect the OOD dataset, enabling the capability of learning unknowns and achieving trustworthy fault diagnosis. Particularly, we develop a probabilistic Bayesian convolutional neural network (CNN) to quantify both epistemic and aleatory uncertainties in fault diagnosis. The fault diagnosis model flags the OOD dataset with large predictive uncertainty for expert intervention and is confident in providing predictions for the data within tolerable uncertainty. This results in trustworthy fault diagnosis and reduces the risk of erroneous decision-making, thus potentially avoiding undesirable consequences. The proposed framework is demonstrated by the fault diagnosis of bearings with three OOD datasets attributed to random number generation, an unknown fault mode, and four common sensor faults, respectively. The results show that the proposed framework is of particular advantage in tackling unknowns and enhancing the trustworthiness of fault diagnosis in safety-critical applications.

---

[*] Corresponding authors.
E-mail address: taotao.zhou@outlook.com (T. Zhou), eald@ucla.edu (E.L. Droguett), mosleh@ucla.edu (A. Mosleh), f.chan@polyu.edu.hk (F. Chan).

# 1. INTRODUCTION

With the proliferation of cheap sensing technology, the increasing connection of internet of things (IoT) devices, and the advances in data analytics, there has been considerable effort in the topic of Prognostics and Health Management (PHM) that utilize multi-sensor data to build maintenance decision support systems to proactively manage asset integrity [1, 2]. Particularly, intelligent health management solutions are developed based on data-driven, model-driven, and hybrid predictive models [3]. An accurate model is often unavailable or difficult to construct [4, 5] due to the complex and dynamic nature of the system's functioning mechanisms and working conditions. Therefore, data-driven models have been dominant in PHM that leverages machine learning techniques [6].

Data-driven models can be categorized based on the data analytic methods adopted being either shallow learning or deep learning method [7]. In general, shallow learning methods are not designed for large-scale datasets and need extensive feature engineering efforts that require large amounts of time and expertise as domain knowledge to manually design features [8]. On the other hand, deep learning methods can handle large-scale datasets and automatically learn from data representing features without deep knowledge about them, allowing the formulation of an end-to-end predictive framework in PHM [9]. As such, deep learning-based PHM has been popularly used for fault diagnosis and remaining useful life (RUL) prognostics. Interested readers are directed to references [10, 11] for a comprehensive review of recent studies of deep learning in the PHM context.

Deep learning-based PHM becomes an emerging solution for end-to-end maintenance decision support systems, especially in the semi- or fully autonomous systems [12, 13], because the inclusion of autonomy raises a critical need for trustworthy diagnosis and prognostics in varying operating environments [14]. However, there are still three main challenges that hinder their deployment in such safety-critical applications:
- Most studies focus on Frequentist deep learning models that represent the weights of a neural network as deterministic values and deliver a single-point estimate. Hence, these models cannot properly convey uncertainty in the predictions [15].
- There is limited literature concerning the vulnerability of deep learning-based PHM to adversarial attacks [16]. Mode and Hoque [17, 18] demonstrated that the RUL prognostics would be greatly compromised by injecting random noise to IoT sensors or by crafting adversarial datasets. Champneys et al. [19] showed the susceptibility of fault diagnosis to adversarial attack by constructing a semantically convincing adversarial dataset to deceive a damage classifier for a three-story structure.
- Most studies implicitly assume that both training and test datasets follow the same distribution [20] and the model performance is validated based on their generalization to the test dataset from the same distribution, referred to as the in-distribution dataset. This implicit assumption is often invalid in reality and the distribution of the test dataset can vary due to various reasons [14].

It is challenging and even infeasible to collect a sufficient amount of annotated training datasets involving the full spectrum of operating environments [21]. The system's behavior can be affected by any unforeseen issue that would cause deviation from the training dataset distribution [14]. Therefore, most deep learning-based PHM is not trustworthy and would make senseless predictions for the dataset not presented in the training dataset, referred to as the out-of-distribution (OOD) dataset [22]. For instance, given irrelevant data like randomly generated multivariate Gaussian noise, the model would still predict healthy or fault conditions even though absurd. This poses challenges for the deployment of deep learning-based predictive models in safety-critical applications.

In the PHM context, the dataset shift can be caused by: (i) the covariate shift between training and testing data, such as the changes of the working condition [23] and sensor faults [24]; (ii) the presence of fault modes that are unseen in the training phase due to the scarcity of failure observations given the high asset reliability [21]; (iii) adversarial attacks could also intentionally cause the dataset shift. In the PHM community, a growing interest is in the utilization of deep domain adaptation and generalization to extrapolate from the training dataset to the test dataset from a different distribution [25, 26]. However, these studies assume some prior knowledge on the connection between the training dataset and test dataset, which may not be reliable or not even available in safety-critical assets [27].

In safety-critical applications, upon receiving OOD data, one should withhold the diagnosis and call for expert intervention instead of relying on the model using implicit assumptions. This is of great importance in safety-critical assets, where unknowns may impose significant risks and cause severe impacts. Therefore, a fault diagnosis model should be able to diagnose faults and also detect the OOD dataset [28]. Herein, we intend to clarify the following concepts:
- Detection aims to identify the existence of any abnormality or fault, while diagnosis focuses on recognizing the exact type of fault [29];
- OOD detection is a closely related area but is different from anomaly detection, outlier detection, or novelty detection, which is typically formulated as a binary classification problem to classify as normal or abnormal [30]. On the other hand, OOD detection deals with multiclass problems. The OOD model needs to maintain satisfactory performance of the intended multiclass classification while also accurately detecting the dataset that is different from the distribution of the training dataset [30, 31, 32].

In this paper, we propose an uncertainty-informed framework for trustworthy fault diagnosis based on the probabilistic Bayesian convolutional neural network (PBCNN). Particularly, the predictive uncertainty conveyed by PBCNN is leveraged to detect the OOD dataset and further diagnose the exact fault type. Our contributions are mainly twofold:

1) Develop a PBCNN-based fault diagnosis framework, which allows one to convey predictive uncertainty considering both epistemic and aleatory uncertainties in fault diagnosis. Particularly, the epistemic uncertainty is considered by the Bayesian CNN layer, which is created by representing the weights of the Frequentist convolution and pooling operations in the form of a probability distribution using the Flipout method. The aleatory uncertainty is considered by a probabilistic output of fault classification following the one-hot categorical distribution. The distribution is parameterized by the output produced by a dense-Flipout layer, which is an extension of the Frequentist dense layer using the Flipout method. Further, the regularization effect is introduced by specifying prior distribution for the weights of the neural network, and the probabilistic output enables one to handle the heterogeneous noise.

2) Propose an uncertainty-informed scheme to enable the fault diagnosis model to learn unknowns not presented in the training phase. Particularly, we examine the effectiveness of various uncertainty measures to express the average or class-wise information of the predictive distribution for OOD detection. Ultimately, we propose to determine the uncertainty threshold by a risk-coverage tradeoff curve to identify OOD datasets without prior knowledge of OOD scenarios. In doing so, in case of large predictive uncertainty, expert interventions can be triggered to avoid erroneous decision-making.

The proposed framework is of particular advantage in safety-critical applications mainly from the following perspectives:

- Avoiding overly confident decisions by quantifying the confidence of fault diagnosis in light of the uncertainties from both model and data sources.
- Enabling the capability of discovering unseen fault mechanisms, which would impose significant risks and potentially cause severe consequences.
- Enhancing the system safety by resorting to human experts when encountering the difficult tasks of fault diagnosis that the model is not confident in predicting.
- Facilitating the understanding of the data quality that may potentially avoid mistakes of data acquisition and annotation.
- Forming an uncertainty-informed approach for the decision-makers to better prioritize the human expert's work orders of fault diagnosis.

The proposed framework is demonstrated using the open-access bearing dataset involving five working conditions, available from Ottawa University. Particularly, we design the numerical experiment by assuming one of the fault modes (i.e., ball defect) is unknown in the training phase and is considered as an OOD dataset to test model performance. Then we train the PBCNN model based on the remaining dataset involving one health mode and three fault modes, in which a test dataset is held out and used as the in-distribution dataset. Three mixed datasets are established to assess model performance, and sensitivity analyses are conducted to validate the proposed framework as follows: (i) a mix of the in-distribution dataset and the OOD dataset randomly generated by the uniform distribution within the range [-1, 1]; (ii) a mix of the in-distribution dataset and the dataset regarding the unknown fault mode of ball defect; (iii) a mix of the in-distribution dataset and the dataset corrupted by four common sensor faults including bias fault, drift fault, scaling fault and precision degradation. Overall, the results validate the effectiveness of the proposed framework in OOD detection and fault diagnosis.

The remainder of the paper is organized as follows. Section 2 summarizes the background of OOD detection, Bayesian deep learning, and a survey on its applications in PHM. Section 3 presents the proposed uncertainty-informed framework for trustworthy fault diagnosis based on PBCNN. Section 4 demonstrates the proposed framework using the Ottawa bearing dataset and provides an extensive evaluation of the proposed framework in dealing with the irrelevant dataset, unknown fault mode, and sensor faults. Section 5 provides concluding remarks and future directions.

## 2. BACKGROUND

### 2.1. Out-of-Distribution Detection

The OOD detection methods can generally be partitioned into two categories [33, 34]. First, develop additional inference mechanisms to detect OOD datasets based on the output feature-space of a pre-trained model. The inference mechanisms are mainly built based on the following motivations: the softmax scores assigned to OOD and in-distribution dataset follows a different distribution [35, 36]; the distance across the feature space of multiple layers within the network would vary between OOD and in-distribution dataset [37]; the probability of OOD dataset is more extreme than the in-distribution dataset according to the extreme value theory [38]. Second, develop a robust representation of the in-distribution dataset and hence be more resilient to the OOD dataset. This is achieved by modifying the network architecture and training process. For instance, the distribution of the in-distribution dataset can be learned by the deep generative model which could output a large reconstruction error given the OOD dataset [39]. Growing attention has recently been given to detect the OOD dataset with the large predictive uncertainty by the bootstrap neural network [40], deep ensembles [41], and Bayesian deep learning [42], among which the latter is the most commonly used [43, 44, 45].

## 2.2. Bayesian Deep Learning

The Bayesian method provides the principal approach in modeling the epistemic uncertainty of deep learning, resulting in Bayesian deep learning such as Bayesian convolutional neural networks [46, 47], and Bayesian Recurrent neural networks [48]. The key of Bayesian deep learning is to model the weight parameter as a probabilistic distribution instead of a determinist value, and ultimately quantify and propagate the epistemic uncertainty [49]. The posterior distributions of the model weights are learned through methods of Bayesian inferences: Markov Chain Monte Carlo (MCMC) and its variants [50]; the variational inference that implicitly exploits model uncertainties through noisy optimization, such as Monte Carlo (MC) Dropout [51, 52]; the variational inference that explicitly models weight parameters as probability distributions such as the Bayes-by-Backprop [53] that enables one to perform mean-field variational inference by the usual backpropagation algorithm. Note that there is debate as to the validity of MC Dropout being Bayesian [54], which in turn, would make the corresponding predictive models problematic in support of reliable uncertainty quantification for health prognostics and fault diagnosis. Therefore, the variational inference is the most commonly used in which reducing the variance of the gradient estimator has been one of the main concerns [55]. Most notably, the Flipout method is a well-recognized implementation that is scalable and efficient by decorrelating the gradients between different training datasets in a mini-batch [56].

## 2.3. Applications of Bayesian Deep Learning in PHM

The PHM community has paid growing attention to Bayesian deep learning with a main focus on quantifying predictive uncertainty. As shown in Table 1, the current literature can be categorized according to the targeting tasks (i.e., RUL prognostics or fault diagnosis), and the Bayesian inference methods adopted. Some important insights are discussed as follows and are the main motivations of our efforts in this paper:

- The vast majority of literature focuses on RUL prognostics and only considers epistemic uncertainty. Besides modeling epistemic uncertainty using MC dropout in [59, 60, 63], Wang et al. [59] model the aleatory uncertainty using Gaussian distribution parameterized by a two-head output; Kim et al. [60] used a feed-forward neural network to model the aleatory uncertainty by assuming a monotonic decreasing relationship between the aleatory uncertainty and RUL; Li et al. [63] modeled aleatory uncertainty by probabilistic output layer following various types of the probability distribution of RUL. Kraus and Feuerriegel [65] also used a probabilistic output layer to examine different hypotheses on the probability distributions of RUL for aleatory uncertainty, while using vanilla Bayes-by-Backprop to model epistemic uncertainty. Caceres et al. [68] modeled epistemic uncertainty using variational inference with the Flipout method and addressed the aleatory uncertainty by a probabilistic output layer following Gaussian distribution.

- Only a small subset of the literature utilized Bayesian deep learning for fault diagnosis. It is worthwhile noting that all of these references adopted MC dropout to model the epistemic uncertainty with applications to crack detection [69, 70], fault detection of chemical processes [71], and fault detection of a centrifugal water-cooled chiller [72]. Therefore, there are two gaps in the current studies, which constitutes one of the main motivations of this paper: (1) there is a need to consider both aleatory and epistemic uncertainties in support of fault diagnosis; (2) explicit variational inference should be used to model epistemic uncertainty due to the methodological deficiency of MC dropout, as discussed in Section 2.2.
- All the current studies assume that training and test datasets follow the same distribution. In other words, there is no study exploring the value of Bayesian deep learning for handling OOD data.

Table 1: A summary of the literature on the applications of Bayesian deep learning in PHM

| Tasks | Bayesian Inference Methods | Reference |
| --- | --- | --- |
| RUL prognostics | MCMC and its variants | [57] |
| RUL prognostics | Implicit variational inference (i.e., MC Dropout) | [58], [59], [60], [61], [62], [63] |
| RUL prognostics | Explicit variational inference (i.e., mean-field variational inference) | [64], [65], [66], [67], [68] |
| Fault diagnosis | MCMC and its variants | N/A |
| Fault diagnosis | Implicit variational inference (i.e., MC Dropout) | [69], [70], [71], [72] |
| Fault diagnosis | Explicit variational inference (i.e., mean-field variational inference) | N/A |

## 3. UNCERTAINTY-INFORMED FRAMEWORK FOR TRUSTWORTHY FAULT DIAGNOSIS

This section describes the uncertainty-informed framework to perform both fault diagnosis and detect the OOD data. The proposed framework consists of three main parts: (a) PBCNN model to quantify the epistemic uncertainty of neural network parameters and aleatory uncertainty inherent in the data observations; (b) based on multiple samples drawn from a well-trained PBCNN model, characterization of the predictive uncertainty of fault diagnosis; (c) leverage the predictive uncertainty to detect the OOD dataset, and further identify the type of fault. The details of each part are discussed in the following section.

## 3.1. Probabilistic Bayesian Convolutional Neural Network

Suppose that the health condition of a component or system can be captured by a temporal series of sensor data. For the data preprocessing, a sliding window is used to cut the sensor data into small segments, and hence the data is reshaped into many segments and window lengths. The data labels are encoded using a one-hot encoding scheme. As illustrated in Figure 1, a segment of sensor data is transformed into its time-frequency representation. This produces an image that is fed into the Bayesian convolutional layer, which is followed by a flatten layer. Then a dense-Flipout layer is used to produce the output results that parameterize the distribution of the one-hot categorical distribution. The detailed architecture configuration and training are presented in Figure 1.

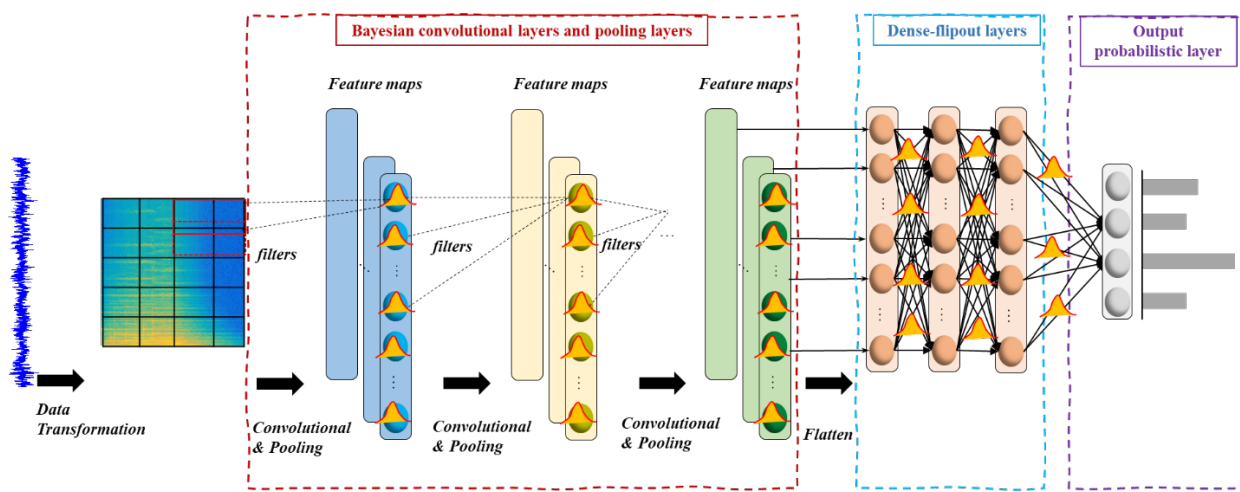

Figure 1: Illustration of the network configuration of the proposed probabilistic Bayesian convolutional neural network.

The epistemic uncertainty is tackled by the Bayesian convolutional layer, which replaces the weights of the Frequentist convolutional layer with the Gaussian distribution using the Flipout method. The Bayesian posterior over the weights would capture the epistemic uncertainty. On the other hand, the aleatory uncertainty is addressed by creating a probabilistic output that follows the one-hot categorical distribution. A dense-Flipout layer is used to produce the parameters of the one-hot categorical distribution in the output layer. As such, the distribution parameters can be learned from the data input, and hence can handle heterogeneous noise within the sensor data. Ultimately, the proposed PBCNN is fully Bayesian and the number of parameters is almost twice its Frequentist counterpart.

Denote the training dataset as $D = \{X, Y\}$, where $X$ represents the images transformed from the sensor measurement and $Y$ represents the one-hot-encoded label representing the working condition. The objective of network training is to learn the variational parameter $\theta$ to parametrize a variational distribution $q(w|\theta)$ that approximates the true posterior distribution of the network weights $p(w|D)$. The objective is to minimize the cost function known as the ELBO [73], which needs a balance between the two terms in Equation (1). The first term measures the complexity cost by the Kullback-Leibler (KL) divergence between the variational distribution $q(w|\theta)$ and the prior distribution of the weights $p(w)$. The second term measures the likelihood cost by calculating the expected value of the likelihood.

$$ELBO(D,\theta) = KL(q(w|\theta)|p(w)) - E_{q(w|\theta)}(\log p(D|w)) \qquad (1)$$

The ELBO can be rewritten in Equation (2) by rearranging the term of complexity cost:

$$\begin{aligned}ELBO(D,\theta) &= E_{q(w|\theta)}(\log q(w|\theta)) - E_{q(w|\theta)}(\log p(w)) \\ &\quad - E_{q(w|\theta)}(\log p(D|w))\end{aligned} \qquad (2)$$

For the sake of computational efficiency, an unbiased estimate of the cost function is derived through Monte Carlo, where $w_i$ denotes the $i^{th}$ Monte Carlo sample drawn from the variational distribution $q(w|\theta)$. The Flipout method is adopted to implement the weight perturbation and the reparameterization trick to generate the Monte Caro samples. This enables the computation of the gradients of variational parameters $\theta$ via backpropagation, to update $\theta$ by gradient-based optimization algorithms:

$$ELBO(D,\theta) \approx \frac{1}{N}\sum_{i=1}^{N} \log q(w_i|\theta) - \log p(w_i) - \log p(D|w_i) \qquad (3)$$

### 3.2. Fault Diagnosis with Uncertainty Quantification

Upon a well-trained PBCNN model, we proceed to predict the fault class given any new data input. For each run of the PBCNN, the fault diagnosis is achieved by a Monte Carlo estimate of the multi-class probabilities by drawing $M$ samples through stochastic forward passes of the network. Accordingly, one can characterize the confidence of the fault diagnosis results based on the mean predictive probability and the predictive uncertainty. The desired property of a confident prediction should have a high mean predictive probability and a small uncertainty.

Denote the predictive probability of the $m^{th}$ sample as a probability vector $p_m = [p_1^m, p_2^m, \dots p_i^m \dots p_N^m]$, where $p_i^m$ is the probability of $i^{th}$ fault class in the $m^{th}$ sample. Then derive a point estimate of the probability regarding each fault class $i$ by computing the mean predictive probability over the M samples: $p^* = [p_1^*, p_2^*, \dots p_i^* \dots p_N^*]$, where $p_i^* = \frac{1}{M}\sum_{m=1}^{M} p_i^m$. The predictive uncertainty is characterized by the predictive entropy and total standard deviation as shown in Equations (4) and (5), respectively. A good prediction should have a small predictive entropy or total standard deviation. In the words, a large predictive entropy or total variance would indicate that the prediction is not trustworthy and the data is unknown to the predictive model.

$$H_T = -\sum_{i=1}^{N} p_i^* \log(p_i^*) \tag{4}$$

$$V_T = \sqrt{\sum_{i=1}^{N} \frac{1}{M-1} \sum_{m=1}^{M} [p_i^m - p_i^*]^2} \tag{5}$$

Note that the predictive entropy can only represent the average amount of information contained in the predictive distribution [74, 75]. The desired property of prediction should also have small predictive uncertainty associated with each class, referred to as class-wise uncertainty. Hence, we propose to also measure the statistical dispersion regarding each class, and then use their maximum values to characterize the predictive uncertainty. Two types of dispersion measures are considered in Equations (6) and (7), referred to as class-wise standard deviation max and class-wise range max, respectively:

$$V_c = \max\left\{\sqrt{\frac{1}{M-1} \sum_{m=1}^{M} [p_i^m - p_i^*]^2} : i = 1, \dots N\right\} \tag{6}$$

$$R_c = \max\{\max(p_i^m) - \min(p_i^m) : i = 1, \dots N\} \tag{7}$$

### 3.3. Uncertainty-Informed Scheme

Figure 2 displays a flowchart of the proposed uncertainty-informed scheme. This is motivated by the fact that examples with uncertainty greater than a user-specified threshold, referred to as untrustworthy examples, are detrimental to the model performance. Particularly, there are mainly three possible scenarios of the untrustworthy examples: (1) in-distribution dataset but with low confidence due to data corruption, such as sensor faults; (2) OOD dataset that is irrelevant to the in-distribution dataset; (3) OOD dataset caused by the mechanisms similar to the in-distribution dataset but unknown in the model training phase. The untrustworthy examples should be flagged before diagnosis and be directed to human experts for examination. A detailed investigation would be triggered if the flagged examples are suspicious (i.e., from an unknown fault mode). On the other hand, the fault diagnosis model is only confident in the trustworthy examples with uncertainty less than the threshold and provides the prediction as to the fault class with the highest mean predictive probability. This reduces the risk of making erroneous decisions, which may lead to undesirable consequences for the safety-critical asset.

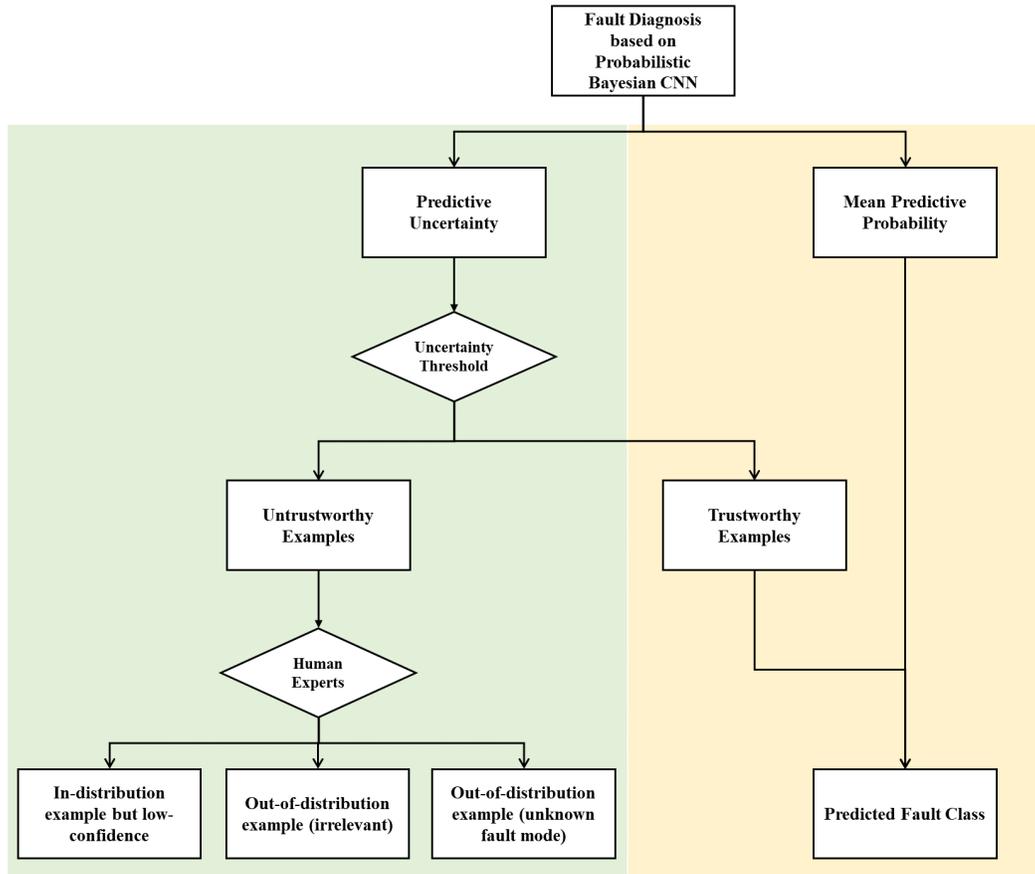

Figure 2: Uncertainty-informed scheme for fault diagnosis in safety-critical applications.

It is essential to specify a proper uncertainty threshold depending on the data confidence, desired performance of OOD detection, and fault diagnosis. A large uncertainty threshold would compromise the model performance and hence pose a higher risk level from two aspects: (i) the model becomes less sensitive to the OOD dataset and incorrectly classifies more OOD examples as in-distribution, leading to poorer performance of OOD detection. (ii) more in-distribution examples are used but adding examples with larger uncertainty may degrade the fault diagnosis performance. Therefore, there are three main questions on the performance assessment of OOD detection and fault diagnosis, and uncertainty threshold specification, which are discussed in the following sections.

### 3.3.1. Performance of OOD Detection

The focus of OOD detection lies in identifying the OOD dataset and meanwhile correctly predicting the in-distribution dataset. Given a specific threshold, the performance of OOD detection can be evaluated based on the two performance indicators in Equations (8) and (9). Herein, the positive examples are referred to as the examples from the in-distribution dataset, and the negative examples are the examples from the OOD dataset:

$$TPR = \frac{TP}{TP + FN} \qquad (8)$$

where TPR is the true positive rate and is defined as the proportion of those in-distribution examples that are correctly predicted, $TP$ is the number of in-distribution examples correctly predicted, FN is the number of in-distribution examples that are wrongly predicted as OOD.

$$FPR = \frac{FP}{FP + TN} \qquad (9)$$

where FPR is the false positive rate and is defined as the portion of those OOD examples that are incorrectly identified as in-distribution, FP is the number of OOD examples that are wrongly predicted as in-distribution, TN is the number of OOD examples that are correctly identified.

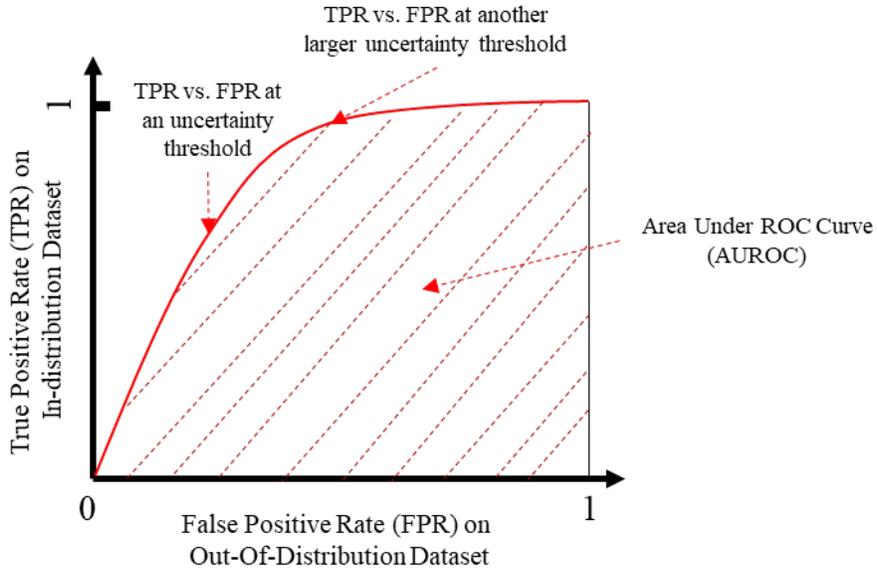

Figure 3: A typical receiver operating characteristic (ROC) curve of detecting in-distribution and out-of-distribution examples at different uncertainty thresholds.

We also evaluate the performance of OOD detection irrespective of the uncertainty threshold based on a threshold-independent metric, that is, the Area Under the Receiver Operating Characteristic curve (AUROC) [35]. AUROC is computed as the area under the ROC curve and is an aggregate measure of OOD detection performance according to the relationship between TPR and FPR. Figure 3 displays a typical ROC curve that can be constructed by evaluating different uncertainty thresholds. Increasing the uncertainty threshold would classify more examples as positive, in turn, increasing both TPR and FPR. Therefore, the ROC curve can be used to assess the usefulness of an uncertainty measure in OOD detection. Additionally, one can evaluate the effectiveness of different uncertainty measures by examining their AUROC measure, which is demonstrated in the following case study. The higher the AUROC, the better the performance of OOD Detection. AUROC of perfect OOD detection is 1.

*3.3.2. Performance of Fault Diagnosis*

The fault diagnosis focuses on identifying the type of faults for the in-distribution dataset. With a certain type of fault condition, we evaluate the performance of fault diagnosis using two performance indicators as below. Herein, the positive examples are referred to as the in-distribution examples under a specific fault condition, and the negative examples are the in-distribution examples under the health condition:

- Precision is defined as the ratio of the number of examples correctly predicted as belonging to a specific fault condition, to the total number of examples predicted as belonging to a specific fault condition.
- Recall is defined as the ratio of the number of examples correctly predicted as belonging to a specific fault condition, to the total number of examples that belong to a specific fault condition.

Given a setup of multi-fault classification involving $G$ types of faults, we aggregate the contributions of each fault condition to compute the average metrics of Precision and Recall [76], which is typically referred to as the micro-average Precision and Recall in Equations (10) and (11). Furthermore, a micro-average F-Measure is used to summarize model performance by being calculated as the harmonic average of the micro-average Precision and Recall in Equation (12). The higher the F-Measure, the better the model performance. An F-Measure of 1 indicates perfect Precision and Recall.

$$Precision_\mu = \frac{\sum_{g=1}^{G} TP_g}{\sum_{g=1}^{G}(TP_g + FP_g^{ID} + FP_g^{OOD})} \qquad (10)$$

where $Precision_\mu$ represents the micro-average Precision, $TP_g$ is the number of examples correctly predicted as belonging to the $g^{th}$ fault condition, $FP_g^{ID}$ is the number of in-distribution examples wrongly predicted as the $g^{th}$ fault condition, $FP_g^{OOD}$ is the number of OOD examples wrongly predicted as the $g^{th}$ fault condition

$$Recall_\mu = \frac{\sum_{g=1}^{G} TP_g}{\sum_{g=1}^{G} (TP_g + FN_g)} \quad (11)$$

where $Recall_\mu$ represents the micro-average Recall, $FN_g$ is the number of examples that belong to the $g^{th}$ fault condition but wrong predicted as the health condition.

$$F_\mu = \frac{2 \times Precision_\mu \times Recall_\mu}{Precision_\mu + Recall_\mu} \quad (12)$$

*3.3.3. Specification of Uncertainty Threshold*

Current practice has no consensus on how to determine the threshold uncertainty. Most studies assume that the OOD dataset is available in the training phase and hence is used to tune the uncertainty threshold. However, this is not true in safety-critical applications due to the lack of prior knowledge of the OOD scenarios. Thus, in order to specify the uncertainty threshold, there is a need for a systemic approach to trace the impact of the uncertainty threshold based on the knowns in the training phase.

We propose to determine the uncertainty threshold based on a trade-off between the data coverage (i.e., the ratio of trustworthy examples with predictive uncertainty lower than a given threshold), and the risk of fault diagnosis model (i.e., the classification error evaluated based on the corresponding trustworthy examples). The objective is to maximize the usage of data (i.e., high data coverage) and meanwhile minimize the predictive risk (i.e., low-risk level). This idea is formally known as the risk-coverage trade-off in the literature [77, 78]. Figure 4 shows a typical risk-coverage curve based on the validation dataset, which can be constructed in three steps:

- Sort the validation dataset into a sequence of ascending uncertainty and calculate the corresponding data coverage as the portion of trustworthy examples in the validation dataset.
- Filter out the examples with larger uncertainty than a given threshold and use the remaining trustworthy examples to evaluate the model performance based on classification error, which indicates the risk of fault diagnosis.
- Establish the relationship between the risk and the data coverage.

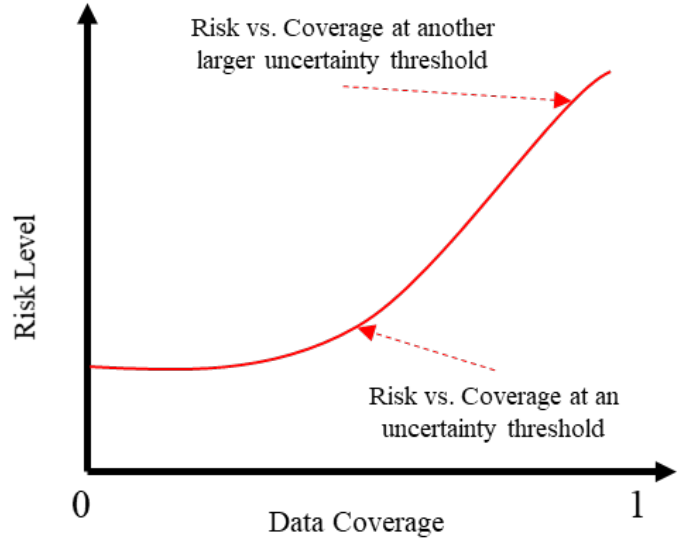

Figure 4: A typical risk-coverage curve to determine the uncertainty threshold.

The risk-coverage curve provides a practical way to trace the impact of the uncertainty threshold without any knowledge of the OOD dataset. Given a specific risk level, the corresponding data coverage and uncertainty threshold can be determined for an uncertainty measure. As such, the risk level can be treated as a hyperparameter to tune the uncertainty threshold, which makes it comparable across different uncertainty measures. Particularly, there are two possible scenarios:

- With a large uncertainty threshold, most of the examples are considered trustworthy for evaluating the model performance and hence a high data coverage. However, such examples have relatively large uncertainty, and adding more such examples would further degrade the model performance and, in turn, increase the predictive risk (i.e., high-risk level).
- Given a small uncertainty threshold, only a small number of examples are trustworthy for evaluating the model performance and hence a low coverage. Such examples have small uncertainty and hence the model is more confident in the predictions with better accuracy and, in turn, in reducing the predictive risk (i.e., low-risk level).

Although the risk level above mainly captures the risk of fault diagnosis on the in-distribution dataset, it can also implicitly reflect the risk of OOD detection because a large uncertainty threshold can not only increase the risk of fault diagnosis but also raise the possibility of classifying more OOD data as in-distribution and hence increasing the risk of OOD detection. As such, the risk level can be viewed as a measure of overall risk in both fault diagnosis and OOD detection.

## 4. CASE STUDY

This section demonstrates the proposed framework using the bearing dataset released by Ottawa University. Section 4.1 provides a brief description of the bearing dataset. Section 4.2 presents the data preparation and model development. Section 4.3 discusses the results using mixed datasets of the in-distribution dataset and three OOD datasets due to randomly generated data, unseen fault mode, and common sensor faults, respectively. We examine the effectiveness of four uncertainty measures in OOD detection and present a sensitivity analysis of model performance under varying uncertainty thresholds to demonstrate the effectiveness of the proposed framework. The model was developed using Python v3.6 [79], TensorFlow v1.13 [80], and TensorFlow Probability v0.6 [81] on a desktop with Intel Core i7 6700 CPU and 32 GB DDR4 RAM.

### 4.1. Problem and Data Description

The Ottawa bearing dataset records the evolution of bearing conditions under time-varying rotational speed conditions and was released by Huang and Baddourat at the University of Ottawa [82, 83]. The bearing has five possible working conditions depending on its fault location: (i) healthy without any defect, (ii) faulty with an inner race defect, (iii) faulty with an outer race defect, (iv) faulty with a ball defect, and (v) faulty with combined defects of the former three locations. The bearing was tested under four different settings of rotational speed: increasing speed, decreasing speed, increasing then decreasing speed, and decreasing then increasing speed. Hence, there are 20 different experimental settings, each of which was replicated 3 times to ensure data authenticity.

The rotational speed data is measured by an encoder with 1,024 cycles per revolution. The vibration data is measured by an accelerometer with a sampling frequency of 200,000 Hz. Note that only the vibration measurement is of interest in this paper regardless of the impacts of rotational speed. As such, the working condition of each experimental replication is captured by vibration data with a sampling duration of 10 seconds. This leads to 60 vibration datasets and each dataset has 2,000,000 data points. Table 2 displays the index of the dataset regarding the replication number given specific rotational speed conditions and health conditions.

Table 2. A summary of the Ottawa bearing dataset.

| Working conditions | | Operating conditions | | | |
|---|---|---|---|---|---|
| Index | Description | Increasing speed | Decreasing speed | Increasing then decreasing speed | Decreasing then increasing speed |
| 0 | Healthy mode | H-A-1; H-A-2; H-A-3 | H-B-1; H-B-2; H-B-3 | H-C-1; H-C-2; H-C-3 | H-D-1; H-D-2; H-D-3 |
| 1 | Fault mode (Inner race defect) | I-A-1; I-A-2; I-A-3 | I-B-1; I-B-2; I-B-3 | I-C-1; I-C-2; I-C-3 | I-D-1; I-D-2; I-D-3 |
| 2 | Fault mode (Outer race defect) | O-A-1; O-A-2; O-A-3 | O-B-1; O-B-2; O-B-3 | O-C-1; O-C-2; O-C-3 | O-D-1; O-D-2; O-D-3 |
| 3 | Fault mode (Ball defect) | B-A-1; B-A-2; B-A-3 | B-B-1; B-B-2; B-B-3 | B-C-1; B-C-2; B-C-3 | B-D-1; B-D-2; B-D-3 |
| 4 | Fault mode (Combined defects on the inner race, the outer race and a ball) | C-A-1; C-A-2; C-A-3 | C-B-1; C-B-2; C-B-3 | C-C-1; C-C-2; C-C-3 | C-D-1; C-D-2; C-D-3 |

## 4.2. Data Preparation and Model Development

### 4.2.1. Data Preparation

The vibration dataset is segmented every 1,024 data points which results in 23,436 segments for each experimental replication. Then, we transform each segment into a spectrogram by the Short-Time Fourier Transform (STFT) with the Hanning window and window length as 6. As illustrated in Figure 5, the time duration of a segment is 0.00512 seconds, and its time-frequency representation is a 33x33 image. Ultimately, the vibration dataset of each experiment replication is transformed into a batch of images of shape [23436, 33, 33, 1]. The pixel value of each image is scaled into the range [-1, 1]. The data labels are one-hot encoded using a four-element binary vector.

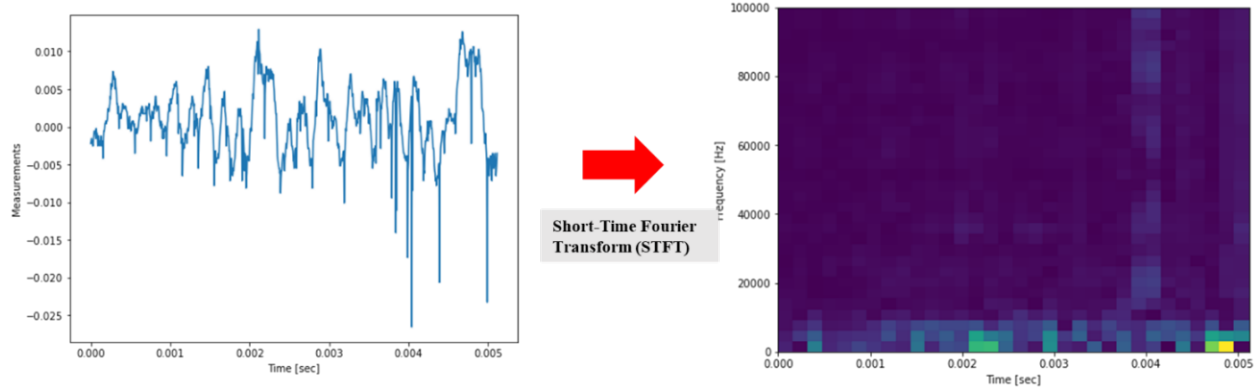

Figure 5: Transform a segment of vibration dataset into an image using the Short-Time Fourier Transform.

Suppose we only know the health mode and the fault modes regarding the inner race defect, outer race defect, and combined defects. Hence, we only have access to the dataset regarding the healthy mode and the three known fault modes for training, 70% of which is used for the training dataset and the rest for the test dataset. Then, we further split the training dataset into validation and training datasets in a ratio of 70:30. All those training, validation, and test datasets follow the same distribution and are referred to as the in-distribution dataset, where the number of examples is well balanced across the working conditions as shown in Table 3. The test dataset with 28,124 examples is held out and used as the in-distribution dataset. The entire dataset regarding ball defects is held out as one type of OOD dataset with 23,436 examples.

Table 3. The number of examples in each working condition of the training, validation, and test datasets.

| Index of working condition | Number of examples in the training dataset | Number of examples in the validation dataset | Number of examples in the test dataset |
|---|---|---|---|
| 0 | 11,504 | 4,901 | 7,031 |
| 1 | 11,461 | 4,944 | 7,031 |
| 2 | 11,510 | 4,895 | 7,031 |
| 4 | 11,459 | 4,946 | 7,031 |

*4.2.2. Network and Training Configuration*

The PBCNN is configured using the same architecture in the Frequentist CNN (FCNN) models but extending the Frequentist convolutional layer and dense layers based on the Flipout method, namely convFlipout layer, and denseFlipout layer. Specifically, the PBCNN architecture consists of three stages. The first stage has two convFlipout layers of 32 feature maps with dimensions 3 × 3, followed by a pooling layer of 32 feature maps with dimensions 2 × 2; the second stage has two convFlipout layers of 64 feature maps with dimension 3 × 3, followed by a pooling layer of 64 feature maps with dimensions 2 × 2. In the third stage, there are three layers: two denseFlipout layers of 100 features each and a denseFlipout layer of four features to parametrize the one-hot categorical distribution regarding four classes. The prior in the PBCNN is a mean-field Gaussian distribution with a mean 0 and standard deviation of 1. Once the model is well trained, 100 samples are drawn through forwarding passes to estimate the probability of fault classes with uncertainty. The number of training epochs is 100 using the Adam optimization algorithm with a learning rate of $1 \times 10^{-3}$.

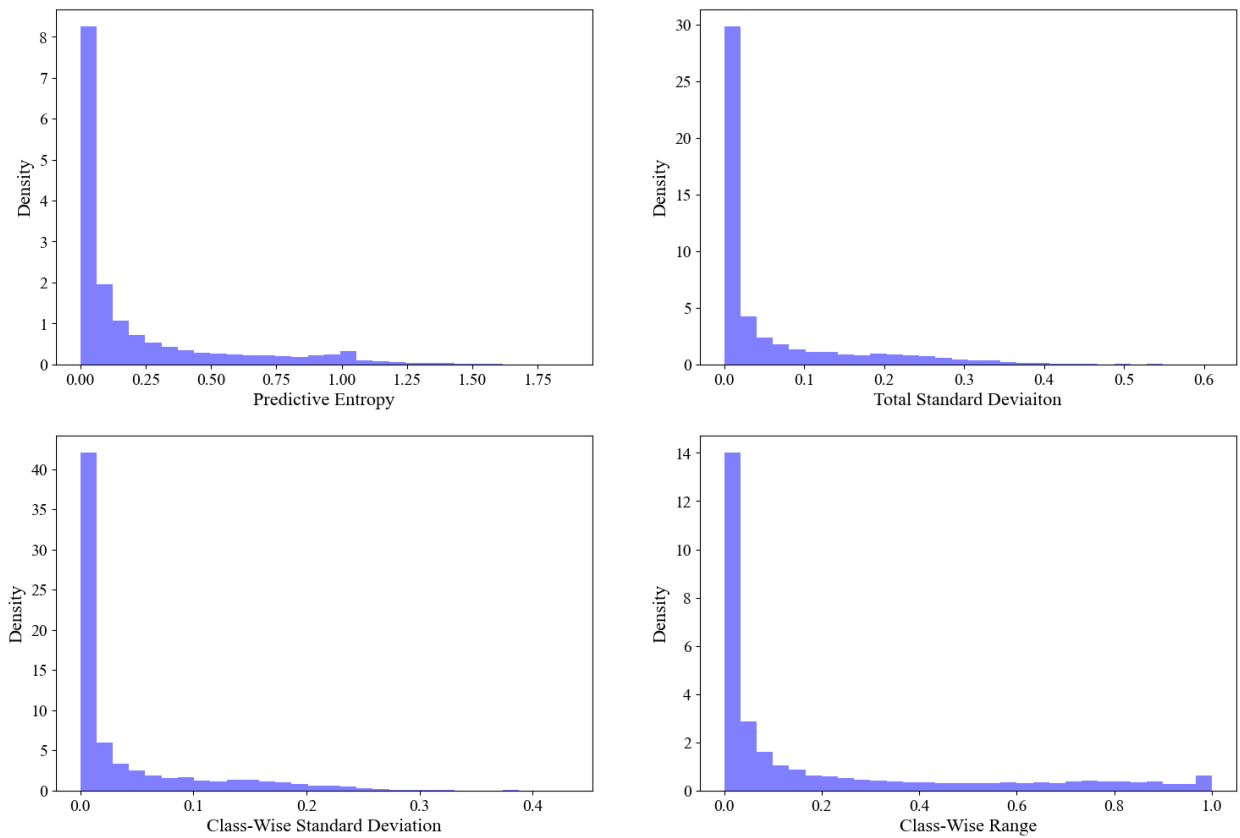

Figure 6: Distribution of predictive uncertainty based on the validation dataset.

*4.2.3. Uncertainty Threshold based on the Validation Dataset*

The uncertainty threshold is specified based on the validation dataset, which has a distribution of predictive uncertainty as summarized in Figure 6. It is expected that most uncertainty measures are close to zero, and the distribution of the uncertainty measures is positively skewed. This provides the basis to construct risk-coverage curves for each of the four uncertainty measures, as shown in Figure 7. As expected, adding more examples with large uncertainty would raise the risk level of fault diagnosis. Additionally, the risk-coverage curves for total standard deviation and class-wise standard deviation are nearly identical. At the same risk level, the risk-coverage curves for the predictive entropy and class-wise range indicate better and worse data coverage, respectively, than the data coverage derived by the other two uncertainty measures.

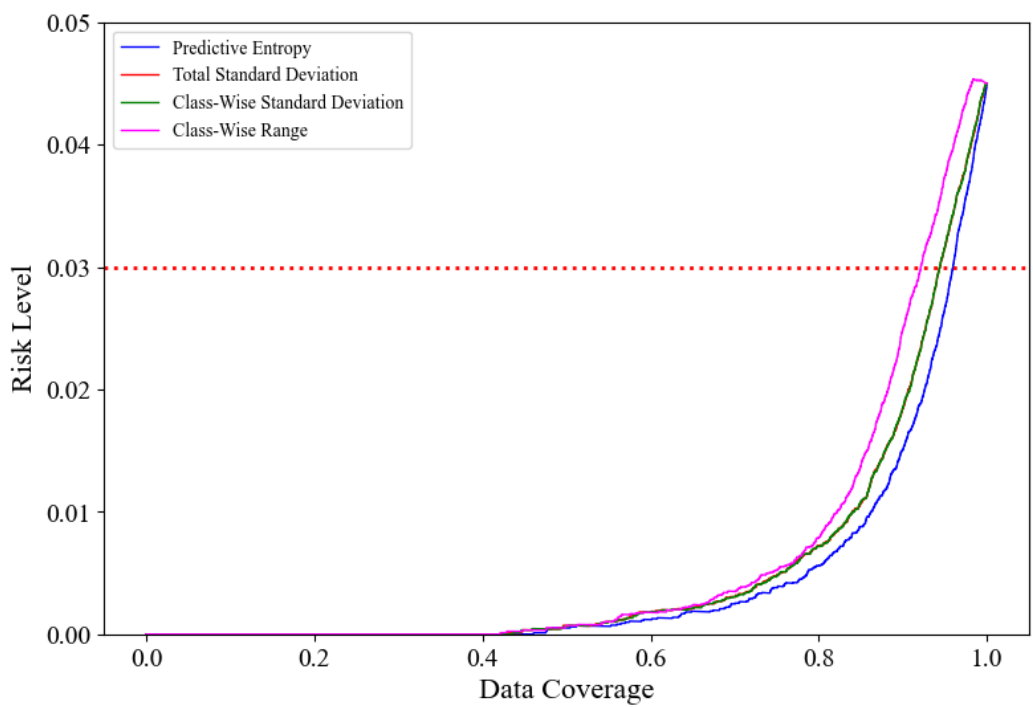

Figure 7: The risk-coverage curve associated with the four types of uncertainty measures.

Suppose the risk level desired by the application is a classification error smaller than 0.03, as illustrated in Figure 7. The threshold value can be specified as 1.005, 0.253, 0.179, and 0.792, for the predictive entropy, total standard deviation, class-wise standard deviation, and class-wise range, respectively. Hence, one can trust the model only in examples where the uncertainty is less than the corresponding threshold. This provides the tools to detect the OOD dataset and low-confidence in-distribution dataset. Furthermore, we also present a sensitivity analysis of the model performance under varying uncertainty thresholds of each uncertainty measure. Table 4 displays the threshold for all four uncertainty measures under 7 different risk levels.

Table 4: Uncertainty threshold of the four uncertainty measures for the sensitivity analysis of risk level.

| Uncertainty Measure | Risk Level | | | | | | |
|---|---|---|---|---|---|---|---|
| | 0.005 | 0.01 | 0.015 | 0.02 | 0.025 | 0.03 | 0.035 |
| Predictive entropy | 0.325 | 0.565 | 0.730 | 0.869 | 0.954 | 1.005 | 1.043 |
| Total standard deviation | 0.068 | 0.137 | 0.178 | 0.208 | 0.231 | 0.253 | 0.282 |
| Class-Wise standard deviation | 0.049 | 0.098 | 0.127 | 0.149 | 0.164 | 0.179 | 0.200 |
| Class-Wise range deviation | 0.270 | 0.494 | 0.611 | 0.687 | 0.738 | 0.792 | 0.849 |

**4.3. Results and Discussions**

This section examines the model performance in dealing with a mix of the in-distribution dataset and some OOD datasets to mimic the real-field applications. Particularly, the model performance is examined in the three mixed datasets as follows:
- A mix of irrelevant data that is randomly generated through the uniform distribution.
- A mix of data regarding an unknown fault mode that is the ball-defect dataset.
- A mix of data corrupted by four common sensor faults including bias fault, drift fault, scaling fault, and precision degradation, respectively.

*4.3.1. A Mix of Irrelevant Data*

This section demonstrates the performance of PBCNN in handling the OOD dataset that is far different from the in-distribution dataset, referred to as the irrelevant dataset. Particularly, we craft the irrelevant dataset of shape [33, 33, 1] by randomly generating its pixel value using the uniform distribution within the range [-1, 1]. The model performance is validated based on a mix of the in-distribution test dataset and the irrelevant dataset consisting of 28,123 irrelevant examples as equivalent to the number of examples in the in-distribution dataset.

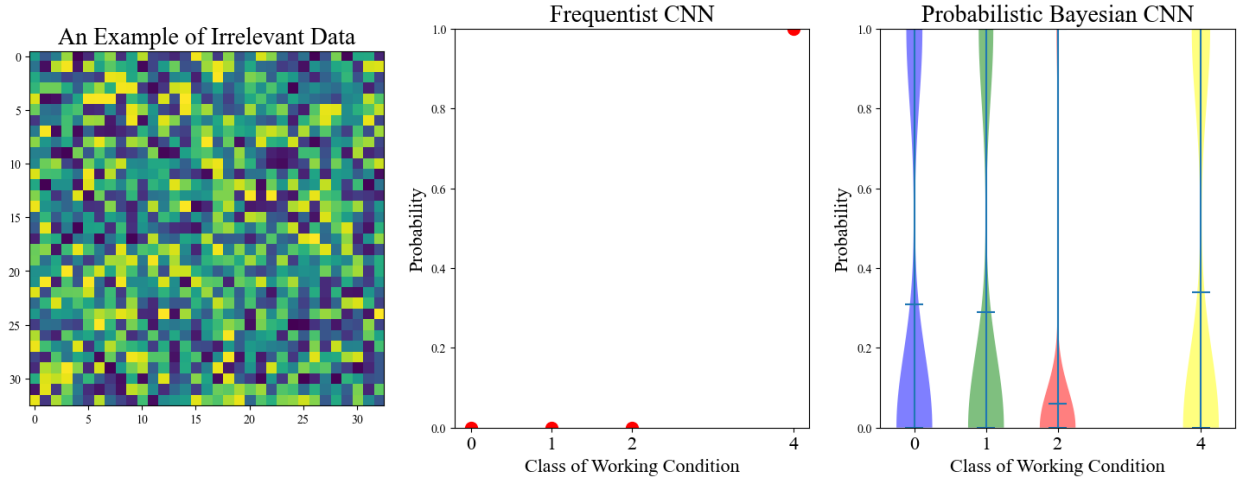

Figure 8: Usage of Frequentist convolutional neural network (CNN) and probabilistic Bayesian CNN for the prediction of an irrelevant example.

To illustrate the unique advantage of PBCNN against FCNN, we present an example of irrelevant data in Figure 8. Some important insights are summarized below:

- FCNN is a deterministic model and can only yield a point estimate of the probability associated with each working condition. However, PBCNN can convey the uncertainty of the probability regarding each working condition, which is visualized using a violin plot.
- FCNN cannot recognize such irrelevant data and yield wrong predictions as any one of the known classes defined through the model training. Given an example of irrelevant data, FCNN tends to provide a high confident prediction (i.e., 1.00) of Class 4. This demonstrates the susceptibility of FCNN in dealing with the OOD dataset, and hence FCNN cannot be trusted in safety-critical applications.
- PBCNN provides a large predictive uncertainty for the probability regarding each working condition. This means the PBCNN model is not confident in its predictions. Hence, the data is flagged as unknown and can be directed to a human expert for further analysis. This shows the advantage of PBCNN in dealing with irrelevant data. Particularly, the predictive entropy, total standard deviation, class-wise standard deviation, and class-wise range are 1.814, 0.837, 0.474, and 1.00, respectively.

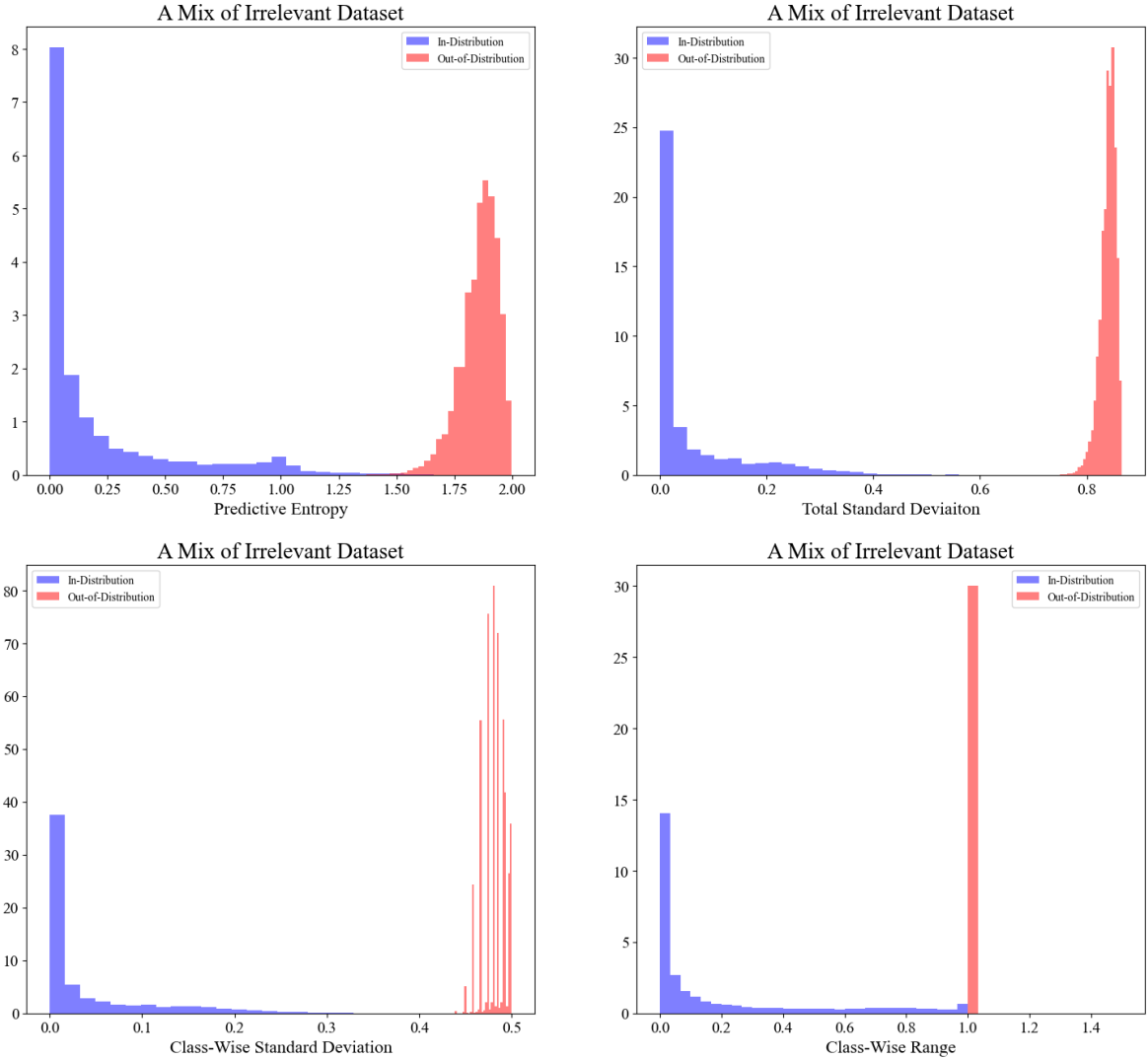

Figure 9: Comparison of the distribution of predictive uncertainty between in-distribution and out-of-distribution datasets due to random noise.

Indeed, random effects exist through data generation and model evaluation. Figure 9 presents the distribution of predictive uncertainty over the entire mixed dataset by comparison between in-distribution and OOD due to the irrelevant dataset. Overall, the OOD dataset has higher predictive uncertainty than the in-distribution dataset. Note that:

- There is a clear difference in the predictive uncertainty between the in-distribution and the OOD dataset for all four uncertainty measures. Hence, setting an appropriate threshold can efficiently distinguish the in-distribution dataset and the irrelevant dataset caused by random number generation.
- It seems that predictive entropy, total standard deviation, and class-wise standard deviation would be generally better than class-wise range because the distance of the predictive uncertainty distribution between in-distribution and out-of-distribution is relatively larger.

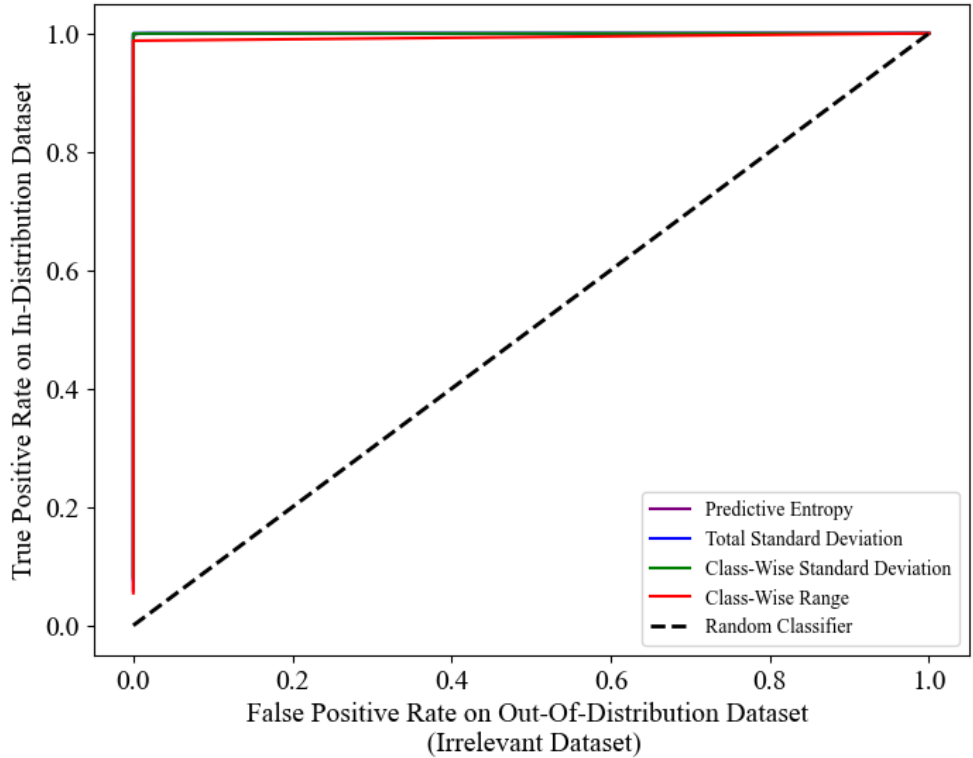

Figure 10: ROC Curve for four uncertainty measures in detecting the irrelevant dataset caused by random number generation.

The performance of the four uncertainty measures in the OOD detection is further examined by their ROC curves in Figure 10. The ROC curve for a random classifier is also shown as a baseline. All four ROC curves are close to the top-left corner and indicate the good performance of all uncertainty measures. Specifically, the AUROC for the predictive entropy, total standard deviation, class-wise standard deviation, and class-wise range are 0.999, 0.999, 0.999, and 0.994, respectively. This corroborates our observations in Figure 9 that the performance of the class-wise range is not as good as the other three uncertainty measures.

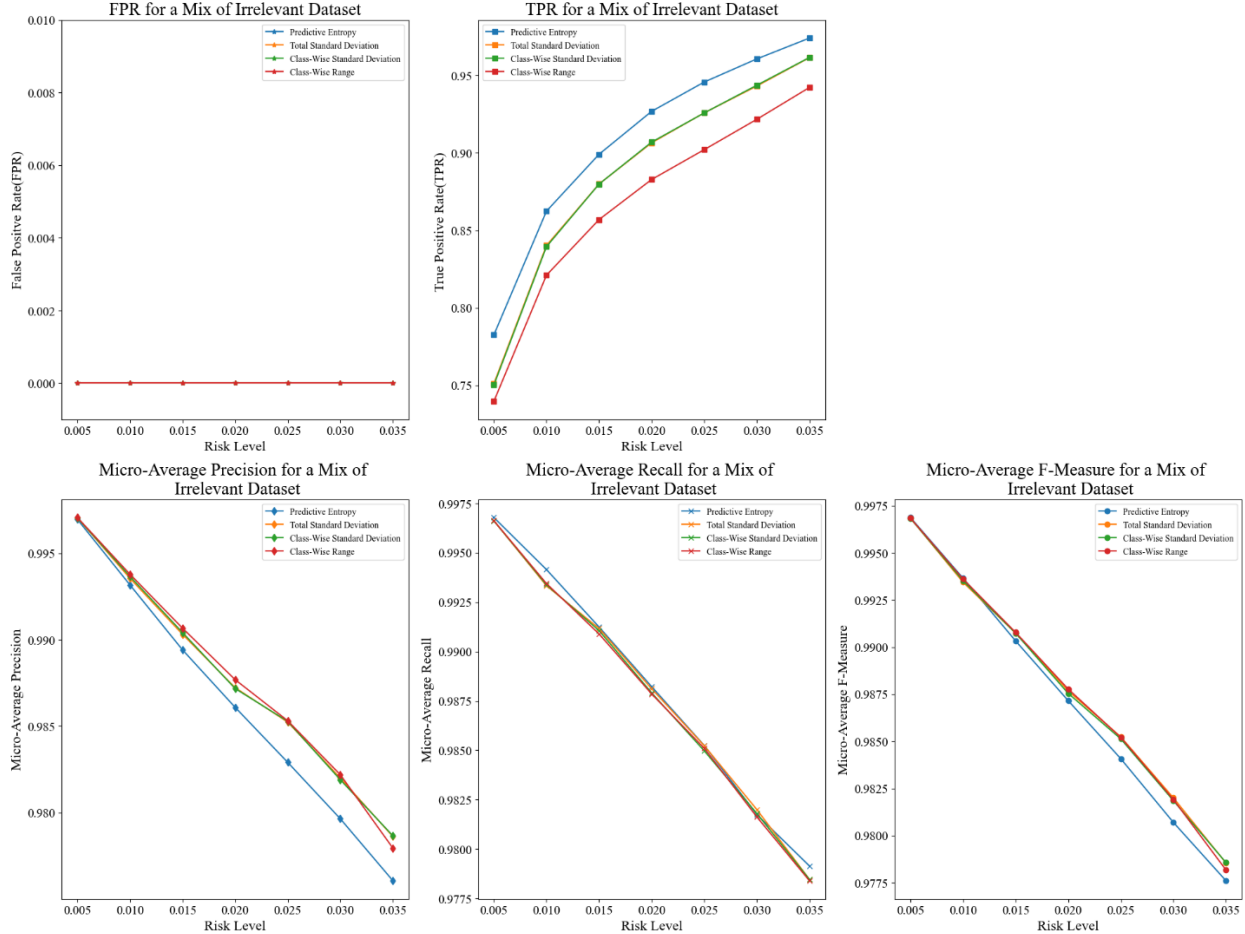

Figure 11: Model performance of four uncertainty measures in OOD detection and fault diagnosis for a mix of the irrelevant dataset under varying risk levels.

We further assess the performance of the proposed framework in OOD detection and fault diagnosis. Particularly, we present a sensitivity analysis of the model performance under varying risk levels or uncertainty thresholds as shown in Table 4. The results are summarized in Figure 11, in which the first row shows the performance of OOD detection, and the second row shows the performance of fault diagnosis:

- Considering the FPR in OOD detection, all four uncertainty measures show a zero FPR and hence correctly identifies all irrelevant examples at all 7 risk levels.
- In fault diagnosis, the performance of total standard deviation and class-wise standard deviation, and class-wise range is close. The predictive entropy is poorer than the other three uncertainty measures because of its lower micro-average Precision at all seven risk levels.
- As discussed in Sections 3.3.1 and 3.3.2, both the TPR of OOD detection and the micro-average Recall of fault diagnosis is only dependent on the risk level and the in-distribution dataset. Hence, the TPR and micro-average Recall are calculated, regardless of the OOD dataset, and remain the same in Sections 4.3.2 and 4.3.3. Specifically, predictive entropy shows a better TPR; total standard deviation and class-wise standard deviation have nearly the same TPR; class-wise range shows the lowest TPR. Additionally, the micro-average Recall of all four uncertainty measures is quite close.

- With the risk level increasing, the performance of fault diagnosis would be compromised by decreasing micro-average Precision, Recall, and F-Measure, which corroborates our claim in Section 3.3.3.

*4.3.2. A Mix of Unknown Fault Mode*

Here we examine the model performance based on the OOD dataset attributed to an unknown fault mode (i.e., ball defect). Figure 12 shows the time-frequency representation of an example of ball defects and the corresponding prediction based on FCNN and PBCNN. Specifically, FCNN would misclassify the example as Class 0 with high confidence. PBCNN would provide a high predictive uncertainty and hence flag the example as unknown for human intervention.

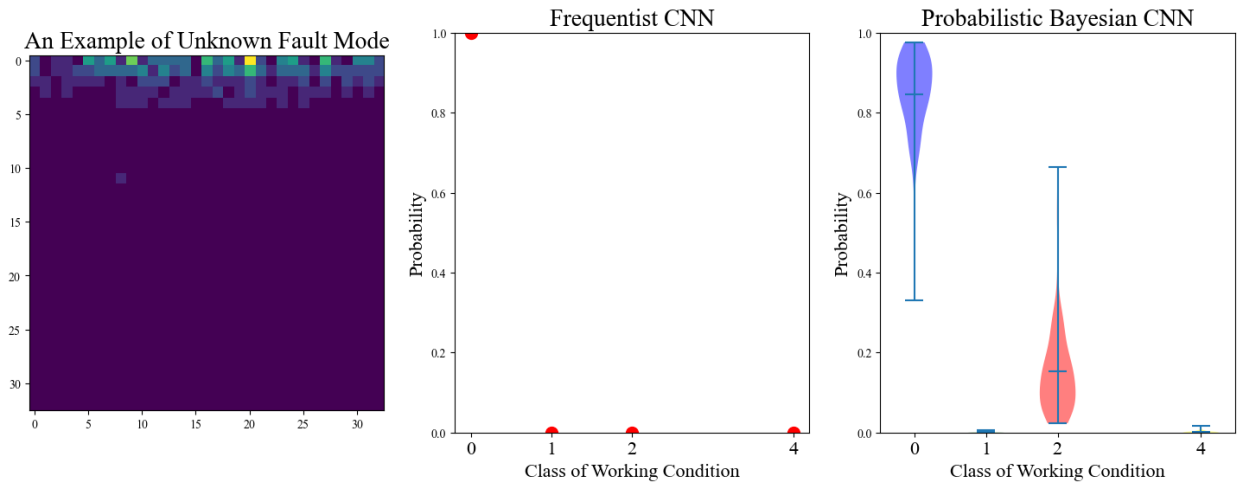

Figure 12: Usage of Frequentist convolutional neural network (CNN) and probabilistic Bayesian CNN for the prediction of an example of unknown fault mode.

Considering the random effects through data features and model evaluation, we examine the distribution of predictive uncertainty for the entire mixed dataset of unknown fault mode in Figure 13. Although most OOD examples have higher predictive uncertainty than in-distribution examples, a high degree of overlap underlies the difficulties in distinguishing the OOD dataset of unknown fault mode, which is discussed in the following sections.

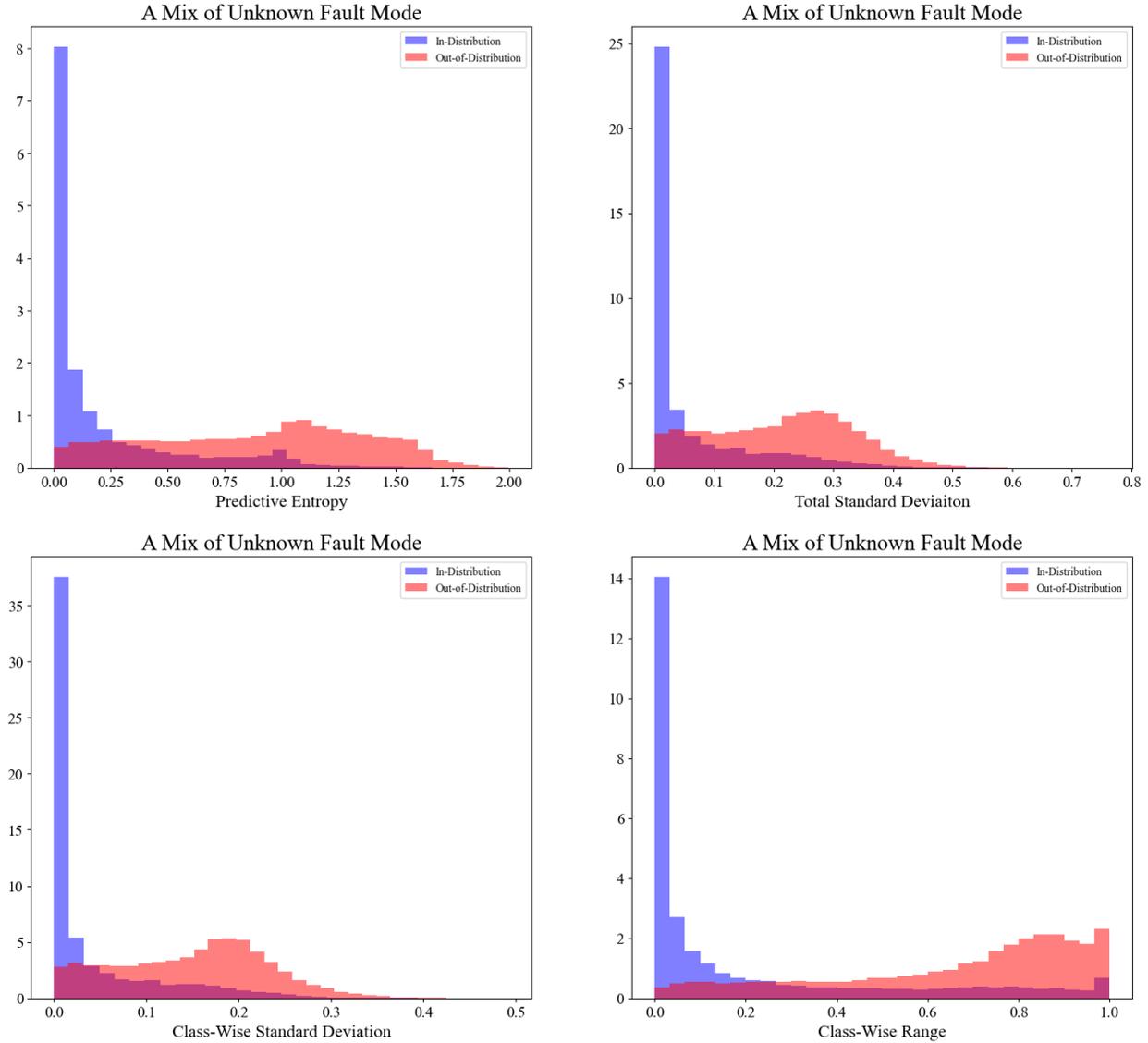

Figure 13: Comparison of the distributions of predictive uncertainty between in-distribution and out-of-distribution datasets due to random noise.

The effectiveness of uncertainty measures in OOD detection is examined by the ROC curves in Figure 14. All uncertainty measures perform well since the ROC curves are all close to the top-left corner. The AUROC for the predictive entropy, total standard deviation, class-wise standard deviation, and class-wise range are 0.895, 0.877, 0.877, and 0.865, respectively. The ROC curves of total standard deviation and class-wise standard deviation almost overlap and hence have nearly the same AUROC. The predictive entropy shows the highest AUROC and the class-wise range has the lowest AUROC. Note that the AUROC values of the four uncertainty measures are similar.

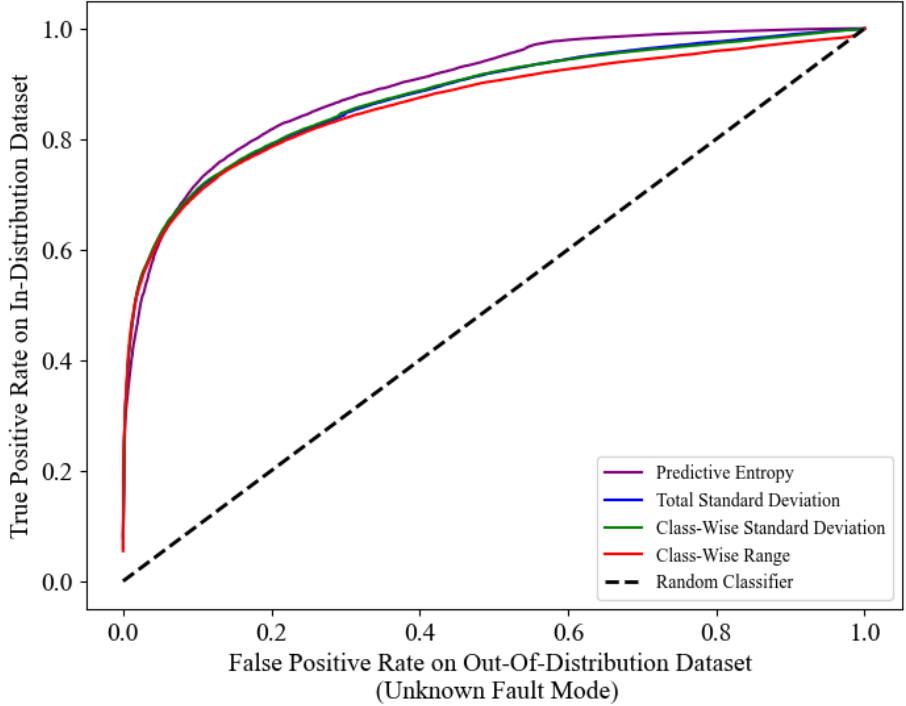

Figure 14: ROC curve for the four uncertainty measures in detecting OOD dataset caused by unknown fault mode.

The results of model performance are displayed in Figure 15, based on the sensitivity analysis of the four uncertainty measures under varying risk levels. As emphasized in 4.3.1, both the TPR and the micro-average Recall are the same as those in Figure 14, regardless of the OOD dataset. Some important insights are discussed below:

- Considering FPR, micro-average Precision and F-Measure, the class-wise range performs the best in both OOD detection and fault diagnosis; the predictive entropy, total standard deviation, and class-wise standard deviation perform similarly.
- It can be noted that the predictive entropy would outperform at a risk level beyond 0.025. This shows that the performance of uncertainty measures would also vary depending on the desired risk level in the specific application.
- When compared to the OOD detection regarding irrelevant dataset, OOD detection of the unknown fault mode becomes more difficult, and further, deteriorates the performance of fault diagnosis.
- As discussed in Section 3.3.3, a higher risk level indicates a worse performance in OOD detection and fault diagnosis, as reflected by the reduced FPR, micro-average Precision and F-Measure.
- Overall, the class-wise range can be recommended given a low-risk tolerance but with slightly compromised TPR; predictive entropy would be used with a higher tolerance for risk.

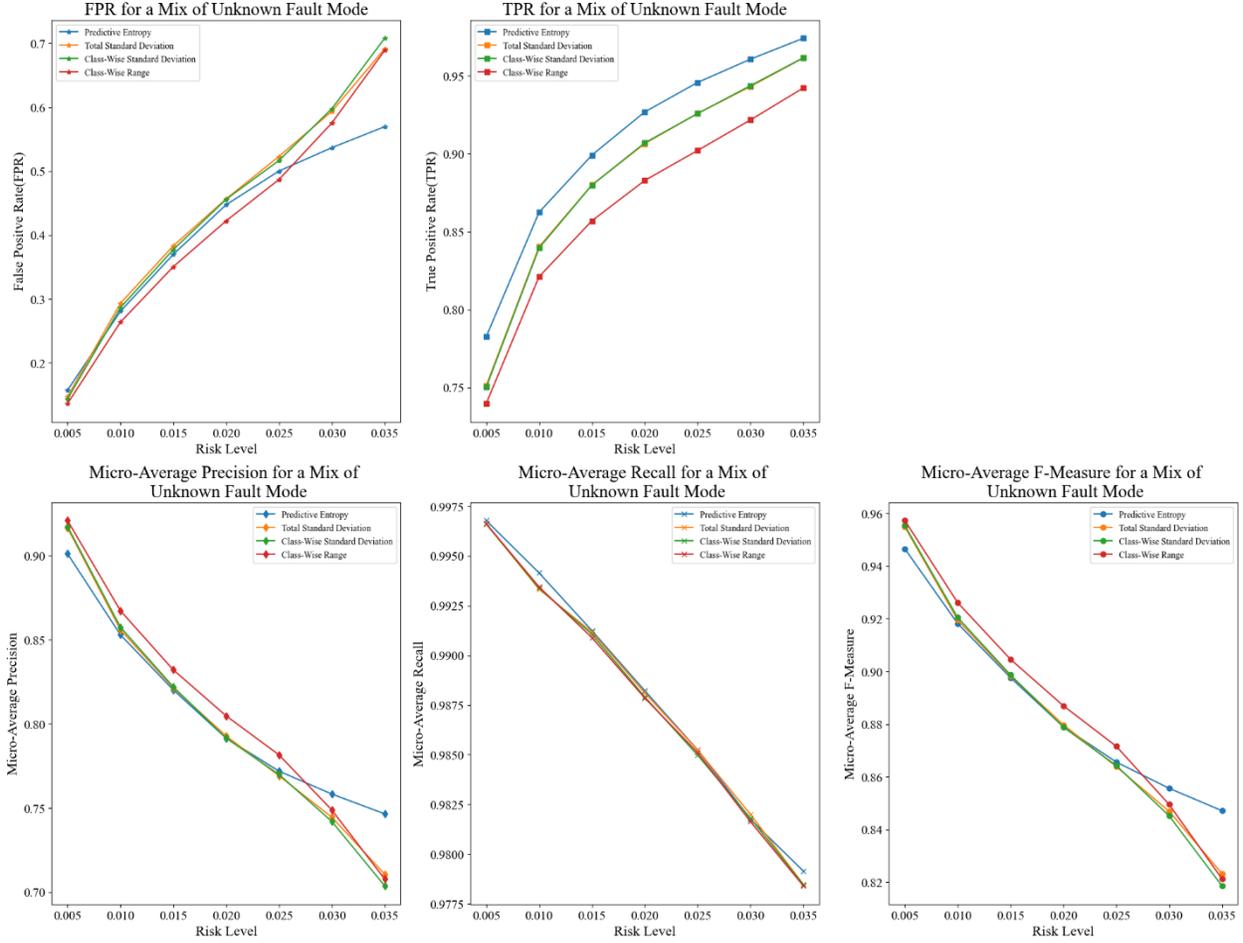

Figure 15: Model performance of four uncertainty measures in OOD detection and fault diagnosis for a mix of unknown fault mode under varying risk levels.

### 4.3.3. A Mix of Sensor Faults

This section examines the model performance in dealing with data corrupted by four common sensor faults [24, 84, 85] bias fault, drifting fault, scaling fault, and precision degradation. In particular, we simulate the fault data by injecting sensor faults into the in-distribution dataset. Note that a sensor fault is assumed to be initiated at the start of a segment. The detailed simulation of a sensor fault is summarized below:

- Bias fault is injected as an offset into a segment of the vibration measurement. The degree of bias $\tau_b$ is specified as a percentage of the peak-to-peak value 0.0753, which is calculated over all segments of the in-distribution dataset. We also introduce Gaussian noise into the bias factor $\tau_b$, with a signal-to-noise ratio (SNR) of 5 dB.
- Drift fault is injected into each segment of the vibration measurement by introducing a linear growing trend. This is specified by a constant drift slope $\tau_d$, which is randomized by Gaussian noise with a SNR of 5 dB.

- Scaling fault is injected by scaling a segment of the vibration measurement. The degree of scaling is specified by a scaling factor $\tau_s$. Gaussian noise is also introduced to $\tau_s$ with a SNR of 5 dB.
- Precision degradation is injected by adding Gaussian noise with zero mean and standard deviation $\tau_p$ as a percentage of the peak-to-peak value. This mimics the behavior of increasing variance due to the precision degradation of a sensor.

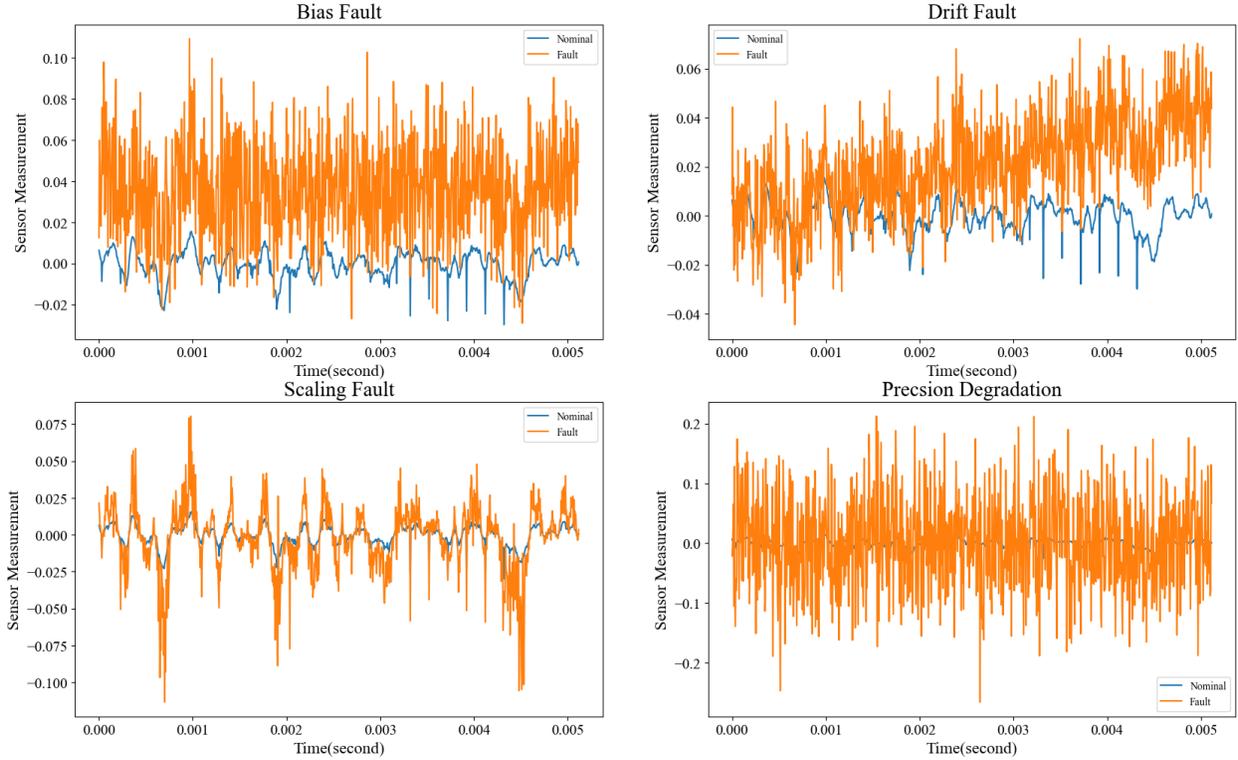

Figure 16: Illustration of fault signal under four common sensor faults including bias, drift, scaling, and precision degradation.

For illustration purposes, we simulate the fault data by setting the parameters as $\tau_b = 0.5$, $\tau_d = 8$, $\tau_s = 3$ and $\tau_p = 1$. An example of the fault data is illustrated in Figure 16. When compared to the nominal sensor measurement, the y-axis value shows a positive offset in the bias fault data; the drift fault data has a linear increasing trend overall; the y-axis value is amplified in the scaling fault data; the precision degradation data has a larger variance. The fault data are then preprocessed and utilized to evaluate the model performance. Figure 17 shows the time-frequency representation of each fault data and their corresponding prediction by FCNN and PBCNN. It is expected that FCNN cannot recognize the corrupted signals and provides overconfident predictions. However, PBCNN reflects large predictive uncertainty in all fault data, indicating that the PBCNN can capture the data corrupted by the four types of sensor faults.

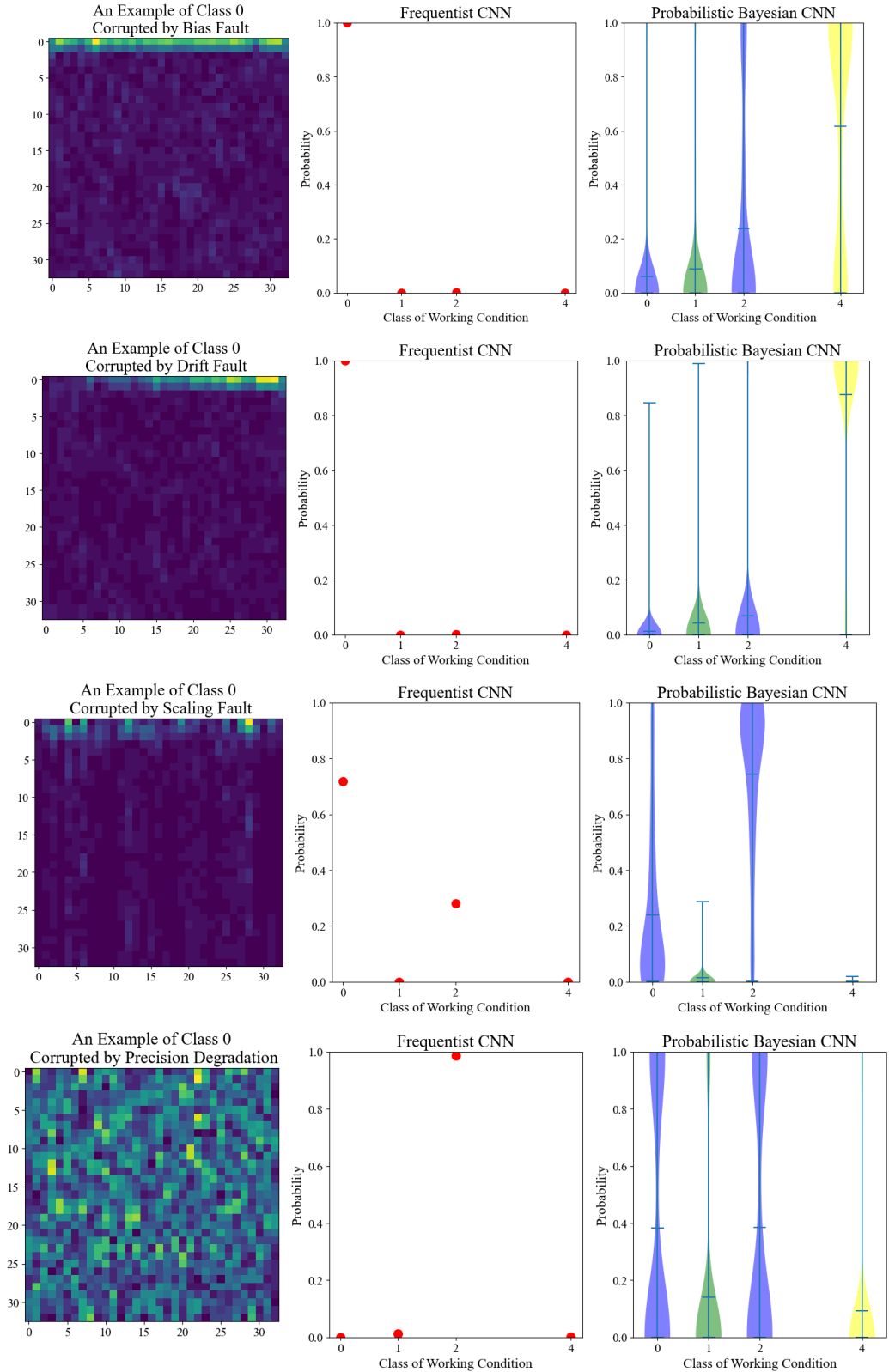

Figure 17: Usage of Frequentist convolutional neural network (CNN) and probabilistic Bayesian CNN for the prediction of the data corrupted by four sensor faults.

We follow the same analysis in Sections 4.3.1 and 4.3.2 to evaluate the performance of the four uncertainty measures in OOD detection. Due to space limitation, the distributions of predictive uncertainty for the four sensor faults are reported in the Appendix. Figure 18 presents the ROC curves for the uncertainty measures in detecting the four sensor faults. The corresponding AUROC results are summarized in Table 5. Some important insights are discussed below:

- All the ROC curves are close to the top-left corner and indicate good performance in detecting sensor faults.
- The sensitivity of the OOD detection varies across different types of sensor faults. Note that the model performs the best in detecting precision degradation, followed by the bias fault, drift fault, and scaling fault.
- Given any type of sensor fault, the performances of uncertainty measures are also different. Specifically, total standard deviation and class-wise standard deviation perform closely; Class-wise range performs the best except a slightly lower AUROC in detecting precision degradation; Predictive entropy is the least effective measure in detecting sensor faults.

Table 5: AUROC of the four uncertainty measures in detection of the four sensor faults.

| Uncertainty Measure \ Sensor Fault | AUROC | | | |
|---|---|---|---|---|
| | Bias Fault | Drift Fault | Scaling Fault | Precision Degradation |
| Predictive entropy | 0.905 | 0.862 | 0.823 | 0.999 |
| Total standard deviation | 0.950 | 0.908 | 0.896 | 0.999 |
| Class-wise standard deviation | 0.950 | 0.908 | 0.896 | 0.999 |
| Class-wise range | 0.953 | 0.915 | 0.907 | 0.994 |

The model performance in OOD detection and fault diagnosis of each sensor fault is also examined by the sensitivity analysis of the uncertainty thresholds corresponding to the 7 risk levels in Table 4. The results are summarized in Figure 19, in which the first column reports the FPR, the second column reports the micro-average Precision, and the third column shows the micro-average F-Measure. Note that the TPR and micro-average Recall are not affected by the type of OOD dataset and remain the same as Sections 4.3.1 and 4.3.2. For the sake of completeness, we report a complete set of metrics regarding each sensor fault in the Appendix. The important insights are summarized below:

- Given any type of fault, the performance ranking of the uncertainty measures is consistent. Specifically, the class-wise range behaves the best, followed by the class-wise standard deviation, total standard deviation, and predictive entropy. Additionally, it is noticed that class-wise range can perform much better than the total standard deviation and class-wise standard deviation.

- Regardless of the uncertainty measure adopted, the model sensitivity is the highest in detecting precision degradation, followed by bias fault, drift fault, and scaling fault.
- Overall, class-wise standard deviation and total standard deviation perform closely, which is consistent with the aforementioned observations. Predictive entropy performs the worst and the class-wise range is the best in OOD detection and fault diagnosis when dealing with sensor faults.
- It is expected that an increased risk level leads to deteriorating performance in both OOD detection and fault diagnosis.
- In light of the recommendations in Section 4.3.1 and 4.3.2, the class-wise range is the most recommended to achieve a balance in handling all the three different OOD datasets.

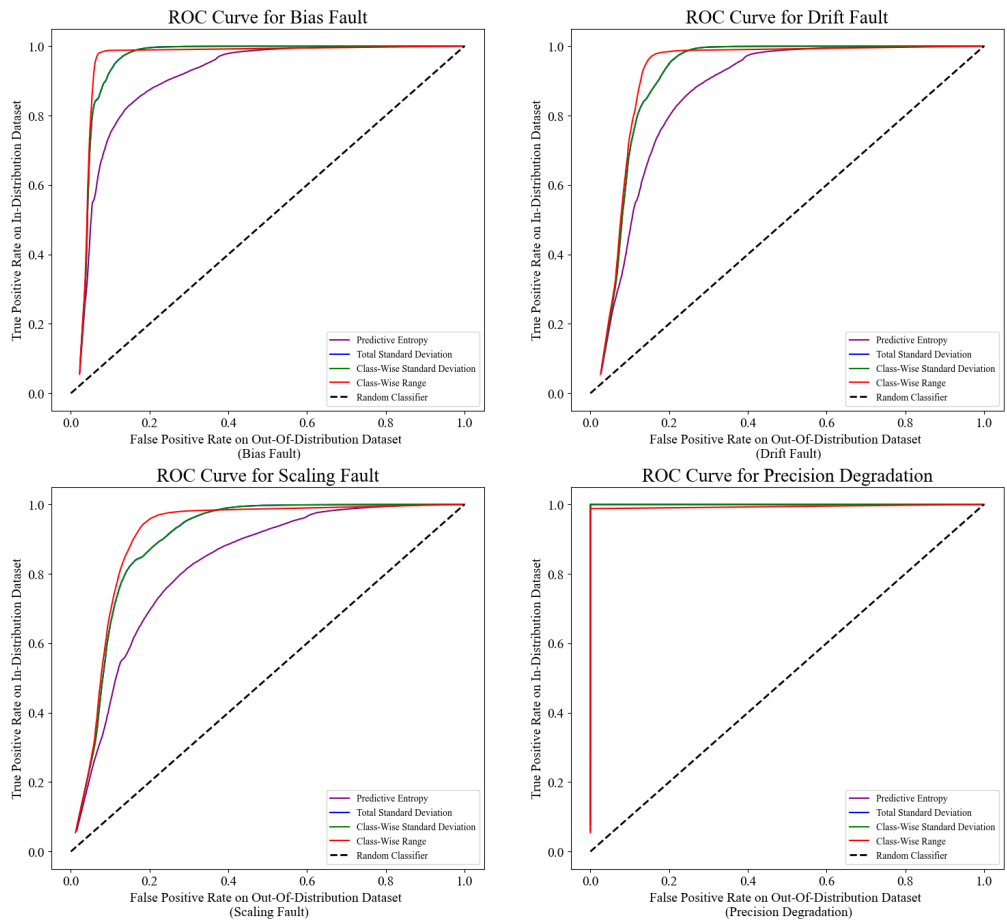

Figure 18: ROC curves for the four uncertainty measures in detecting the OOD dataset due to four sensor faults.

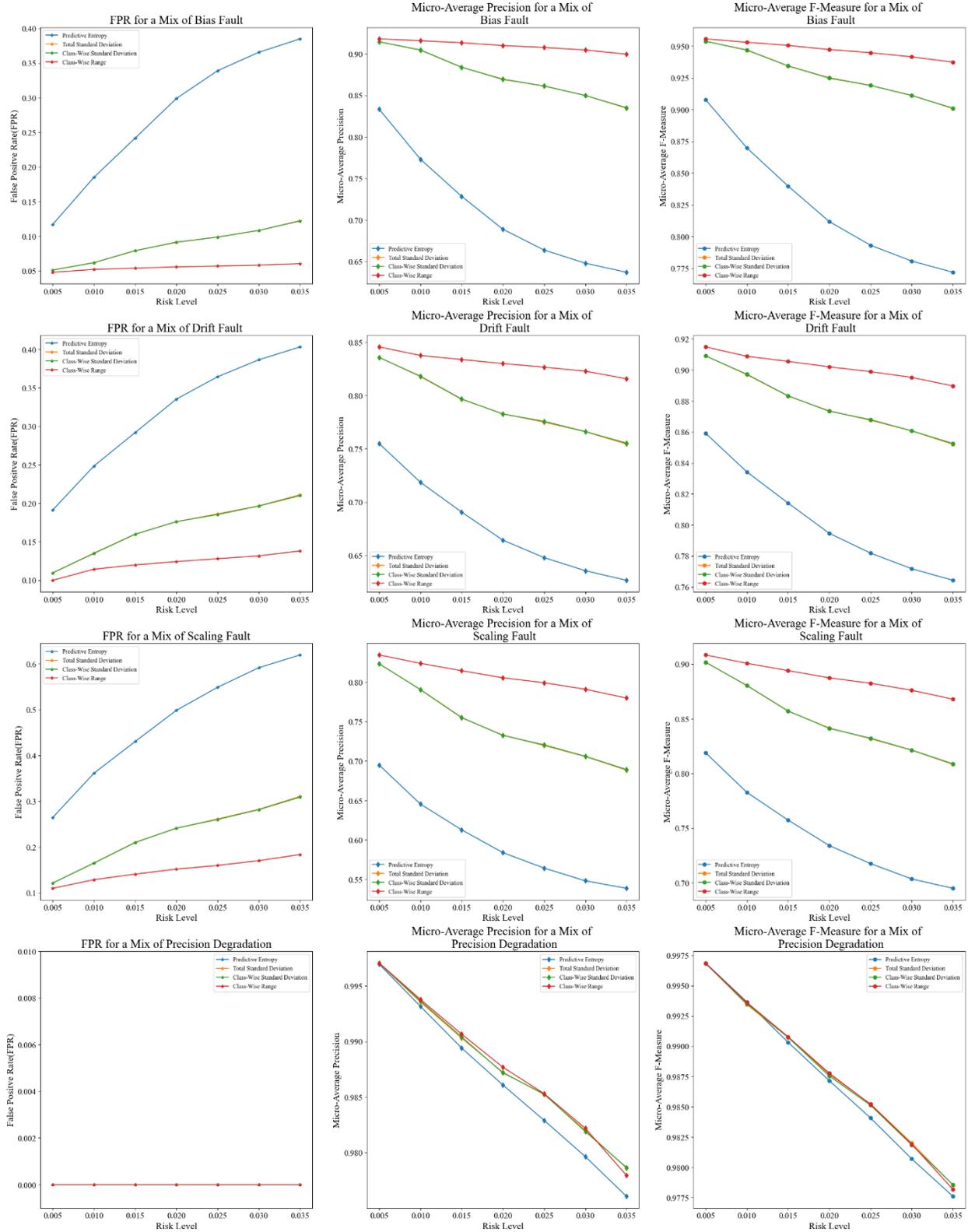

Figure 19: Model performance of four uncertainty measures in OOD detection and fault diagnosis under varying risk levels for a mix of four common sensor faults, respectively.

# 5. CONCLUSIONS AND FUTURE DIRECTIONS

In this paper, we proposed an uncertainty-informed framework to diagnose faults and also detect the OOD dataset by leveraging the predictive uncertainty conveyed by the probabilistic Bayesian convolutional neural network (CNN) considering both epistemic and aleatory uncertainties. This leads to trustworthy fault diagnosis and avoids erroneous decision-making in safety-critical applications by flagging not only the OOD dataset but also the in-distribution dataset with low confidence.

We also demonstrated the application of the proposed framework by the use of an open-access bearing dataset. Particularly, a numerical experiment was designed to examine the model performance involving a mix of the in-distribution dataset and three OOD datasets caused by random number generation, unknown fault model, and four common sensor faults including bias fault, drift fault, scaling fault, and precision degradation. Moreover, it resulted in an extensive evaluation and the model performance according to the TPR and FPR in OOD detection, the micro-average Precision, Recall and F-Measure in fault diagnosis; and on the effectiveness of four uncertainty measures including predictive entropy, total standard deviation, class-wise standard deviation, and class-wise range. The results indicated that the performance of uncertainty measures would vary depending on the type of OOD dataset and the level of risk tolerance. Overall, the class-wise range can be recommended in achieving a balance in terms of performance in OOD detection and fault diagnosis. Moreover, the performance of the total standard deviation and the class-wise standard deviation was close.

The results demonstrated that the proposed framework performs satisfactorily in both OOD detection and fault diagnosis. Future work could encompass benchmarking of both the model performance and computational efficiency using different methods to quantify the uncertainty of deep learning models for fault diagnosis and prognostics such as deep ensemble learning and evidential deep learning; developing more effective mechanisms for characterizing the predictive uncertainty to improve the model performance; conducting more empirical evaluation on the model performance on other types of sensor faults, and the various degree of sensor fault severity.

# REFERENCES

1. Jardine A.K., Lin D., and Banjevic D. "A review on machinery diagnosis and prognostics implementing condition-based maintenance," Mechanical Systems and Signal Processing, 20(7), 1483-1510 (2006).
2. Lei Y., Li N., Guo L., Li N., Yan T., and Lin J. "Machinery health prognostics: A systematic review from data acquisition to RUL prediction," Mechanical Systems and Signal Processing, 104, 799-834 (2018).
3. Liao L., and Köttig F. "Review of hybrid prognostics approaches for remaining useful life prediction of engineered systems, and an application to battery life prediction," IEEE Transactions on Reliability, 63(1), 191-207 (2014).
4. Lee J., Wu F., Zhao W., Ghaffari M., Liao L., and Siegel, D. "Prognostics and health management design for rotary machinery systems—Reviews, methodology and applications," Mechanical Systems and Signal Processing, 42(1-2), 314-334 (2014).
5. Sikorska J.Z., Hodkiewicz M., and Ma L. "Prognostic modelling options for remaining useful life estimation by industry," Mechanical Systems and Signal Processing, 25(5), 1803-1836 (2011).
6. Baraldi P., Cadini F., Mangili F., and Zio E. "Model-based and data-driven prognostics under different available information," Probabilistic Engineering Mechanics, 32, 66-79 (2013).
7. Hamadache M., Jung J.H., Park J., and Youn B.D. "A comprehensive review of artificial intelligence-based approaches for rolling element bearing PHM: shallow and deep learning," JMST Advances, 1(1), 125-151 (2019).
8. Antoni J. and Randall R.B. "The spectral kurtosis: application to the vibratory surveillance and diagnosis of rotating machines," Mechanical Systems and Signal Processing, 20(2), 308-331 (2006).
9. Goodfellow I., Bengio Y., and Courville A. "Deep learning," Cumberland: MIT Press (2016).
10. Jia F., Lei Y., Lin J., Zhou X., and Lu N. "Deep neural networks: A promising tool for fault characteristic mining and intelligent diagnosis of rotating machinery with massive data," Mechanical Systems and Signal Processing, 72-73, pp.303-315 (2016).
11. Zhao R., Yan R., Chen Z., Mao K., Wang P., and Gao R. "Deep learning and its applications to machine health monitoring," Mechanical Systems and Signal Processing, 115, pp.213-237 (2019).
12. Ellefsen A.L., Æsøy V., Ushakov S., and Zhang H. "A comprehensive survey of prognostics and health management based on deep learning for autonomous ships," IEEE Transactions on Reliability, 68(2), 720-740 (2019).
13. Ramos M., Thieme C.A., Utne I.B., Mosleh A., "Proceedings of the 1st International Workshop on Autonomous System Safety (#IWASS)," Trondheim, Norway, March 11-13 (2019).

14. Khan S., Tsutsumi S., Yairi T., and Nakasuka S. "Robustness of AI-based prognostic and systems health management," Annual Reviews in Control, 51, 130-152 (2021).
15. Gal Y. "Uncertainty in deep learning," Doctoral dissertation, University of Cambridge, England (2016).
16. Fink O., Wang Q., Svensen M., Dersin P., Lee W.J., and Ducoffe, M. "Potential, challenges and future directions for deep learning in prognostics and health management applications," Engineering Applications of Artificial Intelligence, 92, 103678 (2020).
17. Mode G.R. and Hoque K.A. "Crafting adversarial examples for deep learning based prognostics," 19th IEEE International Conference on Machine Learning and Applications (ICMLA), Miami, Florida, December 14-17 (2020).
18. Mode G.R., Calyam P., and Hoque K.A. "Impact of false data injection attacks on deep learning enabled predictive analytics," In NOMS 2020-2020 IEEE/IFIP Network Operations and Management Symposium, Budapest, Hungary, April 20-24 (2020).
19. Champneys M.D., Green A., Morales J., Silva M., and Mascarenas D. "On the vulnerability of data-driven structural health monitoring models to adversarial attack," Structural Health Monitoring, 1475921720920233 (2020).
20. Nguyen A., Yosinski J., and Clune J. "Deep neural networks are easily fooled: High confidence predictions for unrecognizable images," In Proceedings of the IEEE conference on computer vision and pattern recognition, Boston, MA, June 7-12 (2015).
21. Jin B. "Incipient anomaly detection with ensemble learning," Doctoral dissertation, University of California, Berkeley (2020).
22. Kenway R. "Vulnerability of deep learning," arXiv preprint arXiv:1803.06111 (2018).
23. Han T., Liu C., Yang W., and Jiang D. "Learning transferable features in deep convolutional neural networks for diagnosing unseen machine conditions," ISA transactions, 93, 341-353 (2019).
24. Balaban E., Saxena A., Bansal P., Goebel K.F., and Curran S. "Modeling, detection, and disambiguation of sensor faults for aerospace applications," IEEE Sensors Journal, 9(12), 1907-1917 (2009).
25. Lu W., Liang B., Cheng Y., Meng D., Yang J., and Zhang T. "Deep model based domain adaptation for fault diagnosis," IEEE Transactions on Industrial Electronics, 64(3), 2296-2305 (2016).
26. da Costa P.R.D.O., Akçay A., Zhang Y., and Kaymak U. "Remaining useful lifetime prediction via deep domain adaptation," Reliability Engineering & System Safety, 195, 106682 (2020).
27. Zhao Z., Zhang Q., Yu X., Sun C., Wang S., Yan R., and Chen X. "Unsupervised deep transfer learning for intelligent fault diagnosis: An open source and comparative study," arXiv preprint arXiv:1912.12528 (2019).
28. Liu W., Wang X., Owens J.D., and Li Y. "Energy-based out-of-distribution detection," arXiv preprint arXiv:2010.03759 (2020).

29. Swain R.R., Dash T., and Khilar P.M. "A complete diagnosis of faulty sensor modules in a wireless sensor network," Ad Hoc Networks, 93, 101924 (2019).
30. Pang G., Shen C., Cao L., and Hengel A.V.D. "Deep learning for anomaly detection: A review," ACM Computing Surveys (CSUR), 54(2), 1-38 (2021).
31. Ren J., Liu P.J., Fertig E., Snoek J., Poplin R., DePristo M.A., Dillon J.V., and Lakshminarayanan B. "Likelihood ratios for out-of-distribution detection," arXiv preprint arXiv:1906.02845(2019).
32. Cao T., Huang C.W., Hui D.Y.T., and Cohen J.P. "A benchmark of medical out of distribution detection," arXiv preprint arXiv:2007.04250 (2020).
33. Pacheco A.G., Sastry C., Trappenberg T., Oore S., and Krohling R.A. "On out-of-distribution detection algorithms with deep neural skin cancer classifiers," In Proceedings of the IEEE/CVF Conference on Computer Vision and Pattern Recognition Workshops, Seattle, WA, June 14-19 (2020).
34. Roady R., Hayes T.L., Kemker R., Gonzales A., and Kanan C. "Are open set classification methods effective on large-scale datasets? " Plos one, 15(9), e0238302 (2020).
35. Hendrycks D., and Gimpel K. "A baseline for detecting misclassified and out-of-distribution examples in neural networks," arXiv preprint arXiv:1610.02136 (2016).
36. Liang S., Li Y., and Srikant R. "Enhancing the reliability of out-of-distribution image detection in neural networks," arXiv preprint arXiv:1706.02690 (2017).
37. Kimin Lee, Kibok Lee, Honglak Lee, and Jinwoo Shin. "A simple unified framework for detecting out-of-distribution samples and adversarial attacks," 32nd Conference on Neural Information Processing Systems (NeurIPS 2018), Montréal, Canada, December 3-8 (2018).
38. Bendale A. and Boult T.E. "Towards open set deep networks," In Proceedings of the IEEE conference on computer vision and pattern recognition, Las Vegas, NV, June 27-30 (2016).
39. Sensoy M., Kaplan L., Cerutti F., and Saleki M. "Uncertainty-aware deep classifiers using generative models," In Proceedings of the AAAI Conference on Artificial Intelligence, New York, February 7-12 (2020).
40. Sedlmeier A., Gabor T., Phan T., Belzner L., and Linnhoff-Popien C. "Uncertainty-based out-of-distribution detection in deep reinforcement learning," arXiv preprint arXiv:1901.02219 (2019).
41. Lakshminarayanan B., Pritzel A., and Blundell C. "Simple and scalable predictive uncertainty estimation using deep ensembles," arXiv preprint arXiv:1612.01474 (2016).
42. Mobiny A., Yuan P., Moulik S.K., Garg N., Wu C.C., and Van Nguyen H. "Dropconnect is effective in modeling uncertainty of Bayesian deep networks," Scientific reports, 11(1), 1-14 (2021).
43. Filos A., Farquhar S., Gomez A.N., Rudner T.G., Kenton Z., Smith L., Alizadeh M., de Kroon A., and Gal Y. "A systematic comparison of Bayesian deep learning robustness in diabetic retinopathy tasks," arXiv preprint arXiv:1912.10481 (2019).

44. Nixon J., Tran D., and Lakshminarayanan B. "Why aren't bootstrapped neural networks better?" 34th Conference on Neural Information Processing Systems (NeurIPS 2020), Virtual-only Conference, December 6-12 (2020).
45. Wilson A.G. " The case for Bayesian deep learning," arXiv preprint arXiv:2001.10995 (2020).
46. Shridhar K., Laumann F., and Liwicki M. "A comprehensive guide to Bayesian convolutional neural network with variational inference," arXiv preprint arXiv:1901.02731 (2019).
47. Gal Y. and Ghahramani Z. "Bayesian convolutional neural networks with Bernoulli approximate variational inference," arXiv preprint arXiv:1506.02158 (2015).
48. Fortunato M., Blundell C., and Vinyals O. "Bayesian recurrent neural networks," arXiv preprint arXiv:1704.02798 (2017).
49. Neal R.M. "Bayesian learning for neural networks (Vol. 118)," Springer Science & Business Media (2012).
50. MacKay D.J. "Information theory, inference and learning algorithms," Cambridge University Press (2003).
51. Salimans T., Kingma D., and Welling M. "Markov chain Monte Carlo and variational inference: Bridging the gap," Proceedings of the 32nd International Conference on Machine Learning, PMLR 37:1218-1226, Lille, France, July 7-9 (2015).
52. Gal Y. and Ghahramani Z. "Dropout as a Bayesian approximation: Representing model uncertainty in deep learning," Proceedings of The 33rd International Conference on Machine Learning, PMLR 48:1050-1059, New York, June 20-22 (2016).
53. Blundell C., Cornebise J., Kavukcuoglu K., and Wierstra D. "Weight uncertainty in neural network," Proceedings of the 32nd International Conference on Machine Learning, PMLR 37:1613-1622, Lille, France, July 7-9 (2015).
54. Osband I., Aslanides J., and Cassirer A. "Randomized prior functions for deep reinforcement learning," 32nd Conference on Neural Information Processing Systems (NeurIPS 2018), Montréal, Canada, December 3-8 (2018).
55. Miller A.C., Foti N.J., D'Amour A., and Adams R.P. "Reducing reparameterization gradient variance," arXiv preprint arXiv:1705.07880 (2017).
56. Wen Y., Vicol P., Ba J., Tran D., and Grosse R. "Flipout: Efficient pseudo-independent weight perturbations on mini-batches," arXiv preprint arXiv:1803.04386 (2018).
57. Benker M., Furtner L., Semm T., and Zaeh M. F. "Utilizing uncertainty information in remaining useful life estimation via Bayesian neural networks and Hamiltonian Monte Carlo," Journal of Manufacturing Systems, (2020).
58. Wei M., Gu H., Ye M., Wang Q., Xu X., and Wu C. "Remaining useful life prediction of lithium-ion batteries based on Monte Carlo Dropout and gated recurrent unit," Energy Reports, 7, 2862-2871 (2021).
59. Wang H., Bai X., and Tan J. "Uncertainty quantification of bearing remaining useful life Based on convolutional neural network," In 2020 IEEE Symposium Series on Computational Intelligence (SSCI), Canberra, ACT, Australia, December 1-4 (2020).

60. Kim M. and Liu K. "A Bayesian deep learning framework for interval estimation of remaining useful life in complex systems by incorporating general degradation characteristics," IISE Transactions, 53(3), 326-340 (2020).
61. Wang B., Lei Y., Yan T., Li N., and Guo, L. "Recurrent convolutional neural network: A new framework for remaining useful life prediction of machinery," Neurocomputing, 379, 117-129 (2020).
62. Peng W., Ye Z.S., and Chen N. "Bayesian deep-learning-based health prognostics toward prognostics uncertainty," IEEE Transactions on Industrial Electronics, 67(3), 2283-2293 (2019).
63. Li G., Yang L., Lee C.G., Wang X., and Rong M. "A Bayesian deep learning RUL framework integrating epistemic and aleatoric uncertainties," IEEE Transactions on Industrial Electronics (2020).
64. Wang R., Chen H., and Guan C. "A Bayesian inference-based approach for performance prognostics towards uncertainty quantification and its applications on the marine diesel engine," ISA transactions (2021).
65. Kraus M. and Feuerriegel S. "Forecasting remaining useful life: Interpretable deep learning approach via variational Bayesian inferences," Decision Support Systems, 125, 113100. (2019).
66. Mazaev T., Crevecoeur G., and Van Hoecke, S. "Bayesian convolutional neural networks for RUL prognostics of solenoid valves with uncertainty estimations," IEEE Transactions on Industrial Informatics (2021).
67. Gao G., Que Z., and Xu Z. "Predicting remaining useful life with uncertainty using recurrent neural process," In 2020 IEEE 20th International Conference on Software Quality, Reliability and Security Companion (QRS-C), Macau, China, December 11-14 (2020).
68. Caceres J., Gonzalez D., Zhou T., and Droguett E.L. "A probabilistic Bayesian recurrent neural network for remaining useful life prognostics considering epistemic and aleatory uncertainties," Structural Control and Health Monitoring, e2811 (2021)
69. Sajedi S.O. and Liang X. "Uncertainty-assisted deep vision structural health monitoring," Computer-Aided Civil and Infrastructure Engineering, 36(2), 126-142 (2021).
70. Bueno de Mesquita A.T. "A Bayesian convolutional neural network approach for image-based crack detection and a maintenance application," Master Thesis, Erasmus University Rotterdam (2021).
71. Sun W., Paiva A.R., Xu P., Sundaram A., and Braatz R.D. "Fault detection and identification using Bayesian recurrent neural networks," Computers & Chemical Engineering, 141, 106991 (2020).
72. Jin B., Li D., Srinivasan S., Ng S. K., Poolla K., and Sangiovanni-Vincentelli A. "Detecting and diagnosing incipient building faults using uncertainty information from deep neural networks," In 2019 IEEE International Conference on Prognostics and Health Management (ICPHM), San Francisco, CA, June 17-20 (2019).

73. San Martin G., Droguett E.L., Meruane V., and das Chagas Moura M. "Deep variational auto-encoders: A promising tool for dimensionality reduction and ball bearing elements fault diagnosis," Structural Health Monitoring, 18(4), 1092-1128 (2019).
74. Kendall A. and Gal Y. "What uncertainties do we need in Bayesian deep learning for computer vision? " arXiv preprint arXiv:1703.04977 (2017).
75. Duerr O., Sick B., and Murina, E. "Probabilistic deep learning: With Python, Keras and TensorFlow Probability," Manning Publications (2020).
76. Frank S.M., Lin G., Jin X., Singla R., Farthing A., Zhang L., and Granderson, J. "Metrics and methods to assess building fault detection and diagnosis tools (No. NREL/TP-5500-72801)," National Renewable Energy Lab (NREL), Golden, CO (2019).
77. El-Yaniv R. "On the foundations of noise-free selective classification," Journal of Machine Learning Research, 11(5) (2010).
78. Zhang J., Kailkhura B., and Han T.Y.J. "Leveraging uncertainty from deep learning for trustworthy material discovery workflows," ACS Omega, 6(19), 12711-12721 (2021).
79. Van Rossum G. and Drake F.L. "Python 3 reference manual," Scotts Valley, CA: CreateSpace (2009).
80. Abadi M. Barham P., Chen J., Chen Z., Davis A., Dean J., Devin M., Ghemawat S., Irving G., Isard M., Kudlur M., Levenberg J., Monga R., Moore S., Murray D.G., Steiner B., Tucker P., Vasudevan V., Warden P., Wicke M., Yu Y., and Zheng X. "Tensorflow: A system for large-scale machine learning," Proceedings of the 12th USENIX Symposium on Operating Systems Design and Implementation (OSDI '16), Savannah, GA, USA, November 2–4 (2016)
81. Dillon J., Langmore I., Tran D., Brevdo E., Vasudevan S., Moore D., Patton B., Alemi A., Hoffman M., and Saurous R. "Tensorflow distributions," arXiv preprint arXiv:1711.10604 (2017).
82. Huang H. and Baddour N. "Bearing vibration data collected under time-varying rotational speed conditions," Data in brief, 21, 1745-1749 (2018).
83. Huang H. and Baddour N. "Bearing vibration data under time-varying rotational speed conditions," Mendeley Data, V2, doi: 10.17632/v43hmbwxpm.2 (2019)
84. Qin S.J. and Li W. "Detection, identification, and reconstruction of faulty sensors with maximized sensitivity," AIChE Journal, 45(9), 1963-1976 (1999).
85. da Silva J.C., Saxena A., Balaban E., and Goebel K. "A knowledge-based system approach for sensor fault modeling, detection and mitigation," Expert Systems with Applications, 39(12), 10977-10989 (2012).

# APPENDIX

## A.1. Bias Fault.

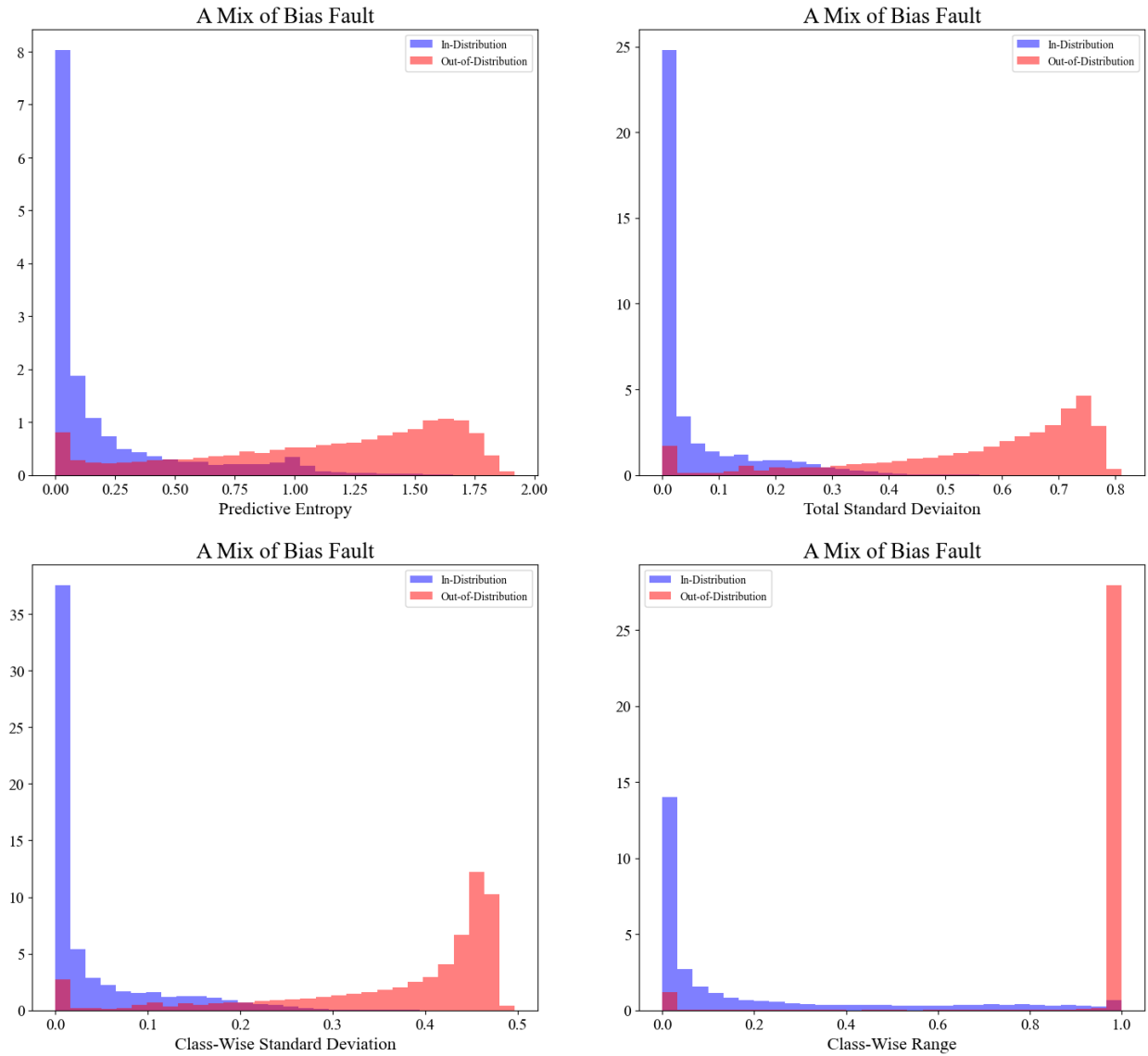

Figure A.1: Comparison of the distributions of predictive uncertainty between in-distribution and out-of-distribution datasets due to bias fault.

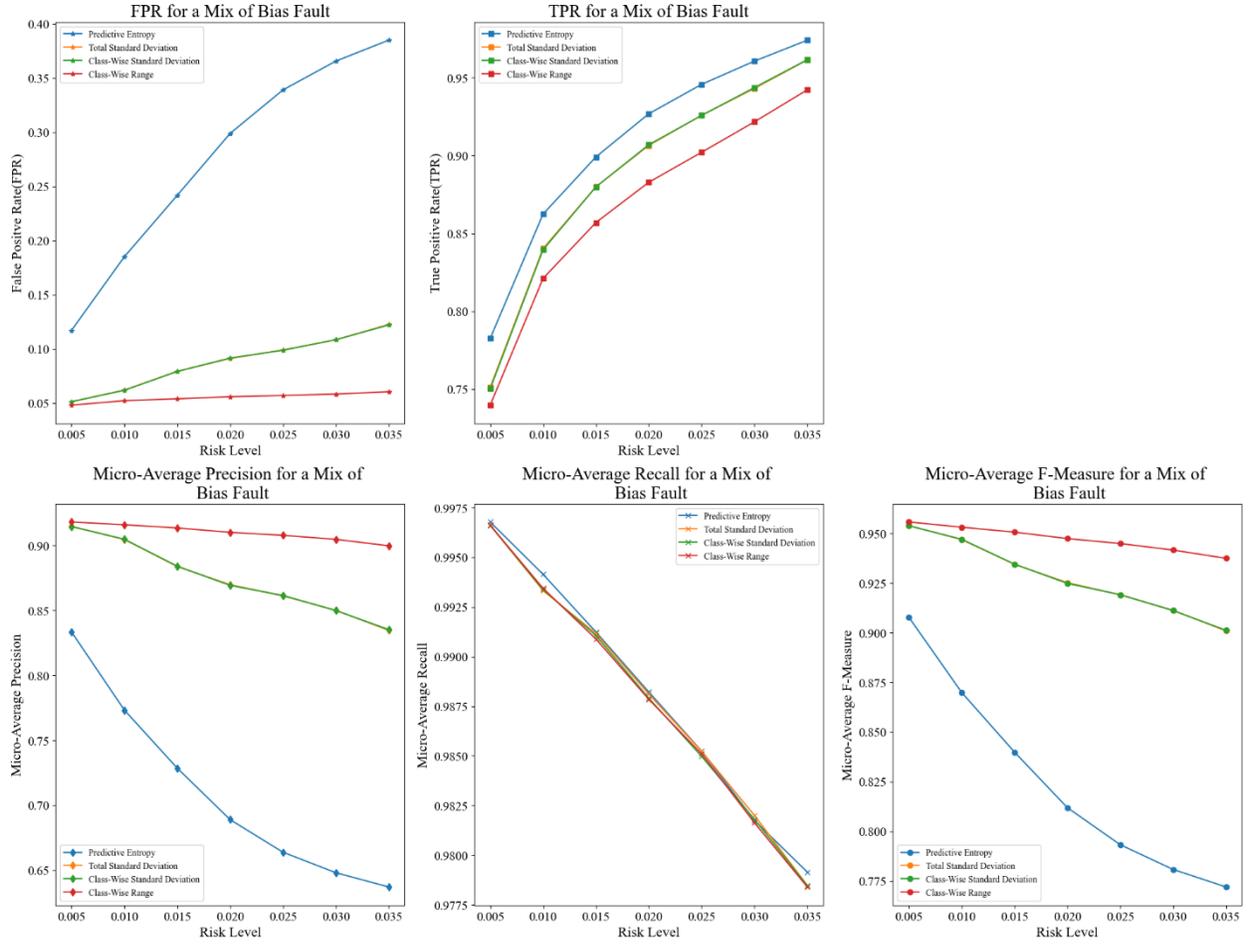

Figure A.2: Model performance of four uncertainty measures in OOD detection and fault diagnosis for a mix of bias fault under varying risk levels.

## A.2. Drift Fault.

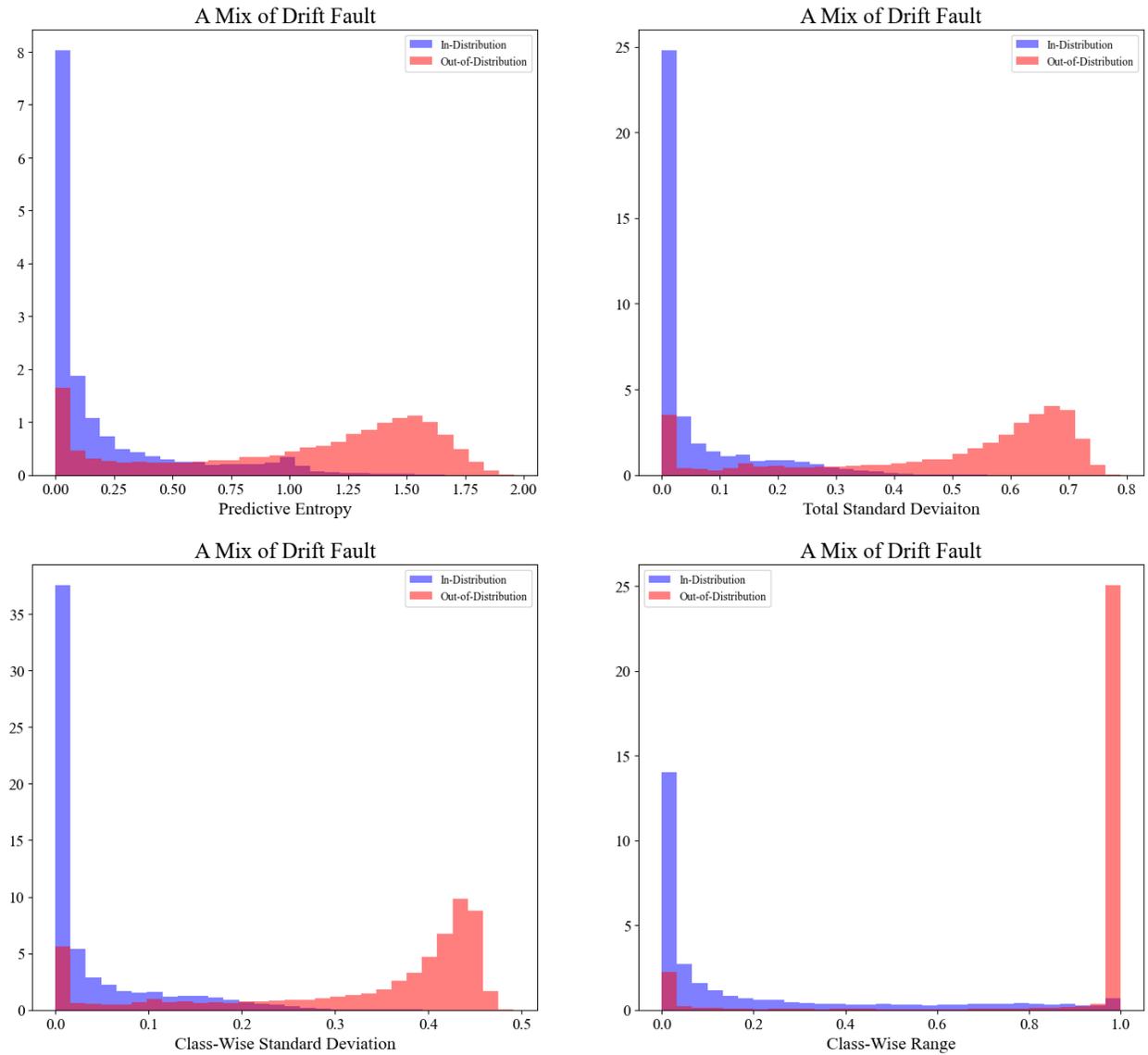

Figure A.3: Comparison of the distributions of predictive uncertainty between in-distribution and out-of-distribution datasets due to drift fault.

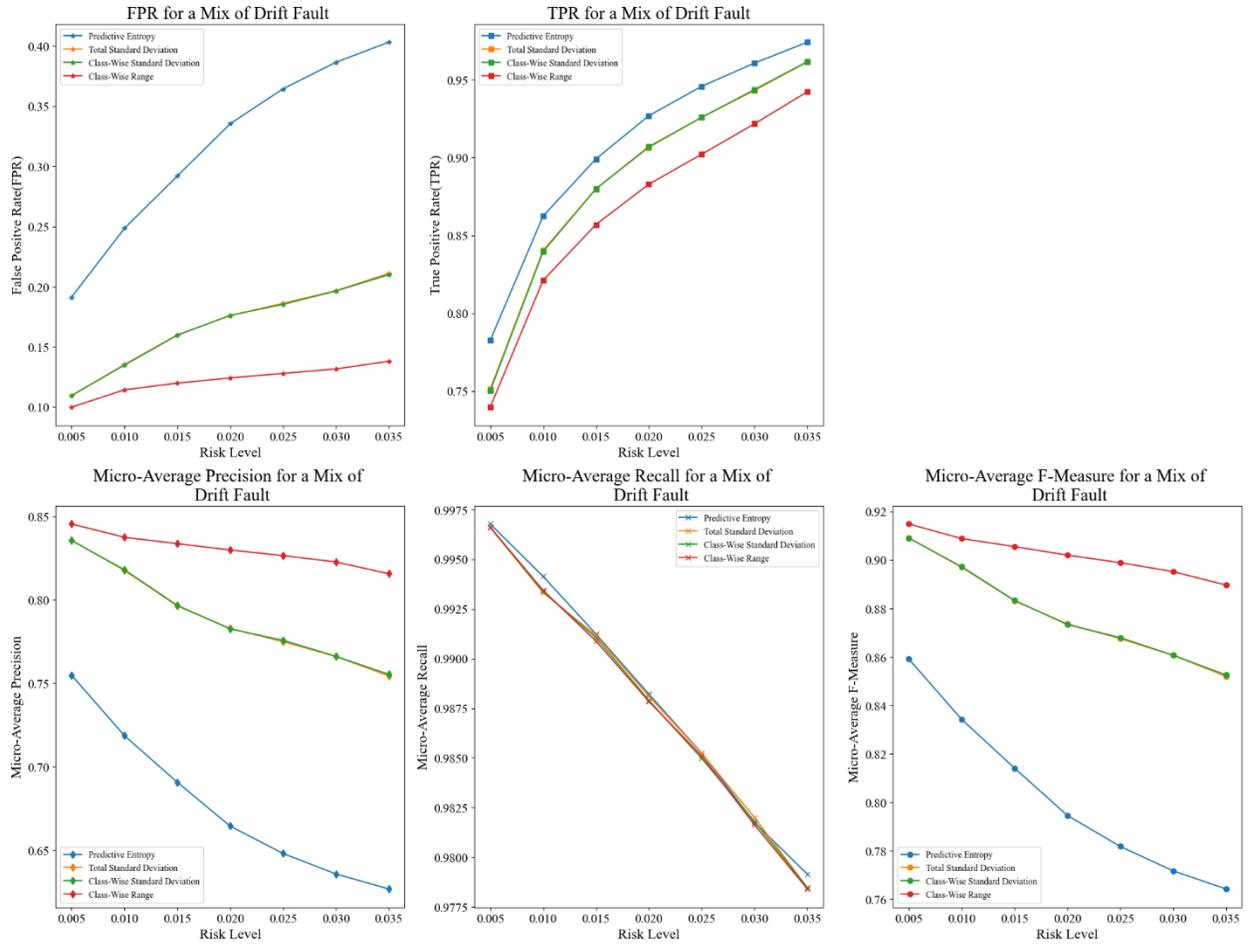

Figure A.4: Model performance of four uncertainty measures in OOD detection and fault diagnosis for a mix of drift fault under varying risk levels.

## A.3. Scaling Fault.

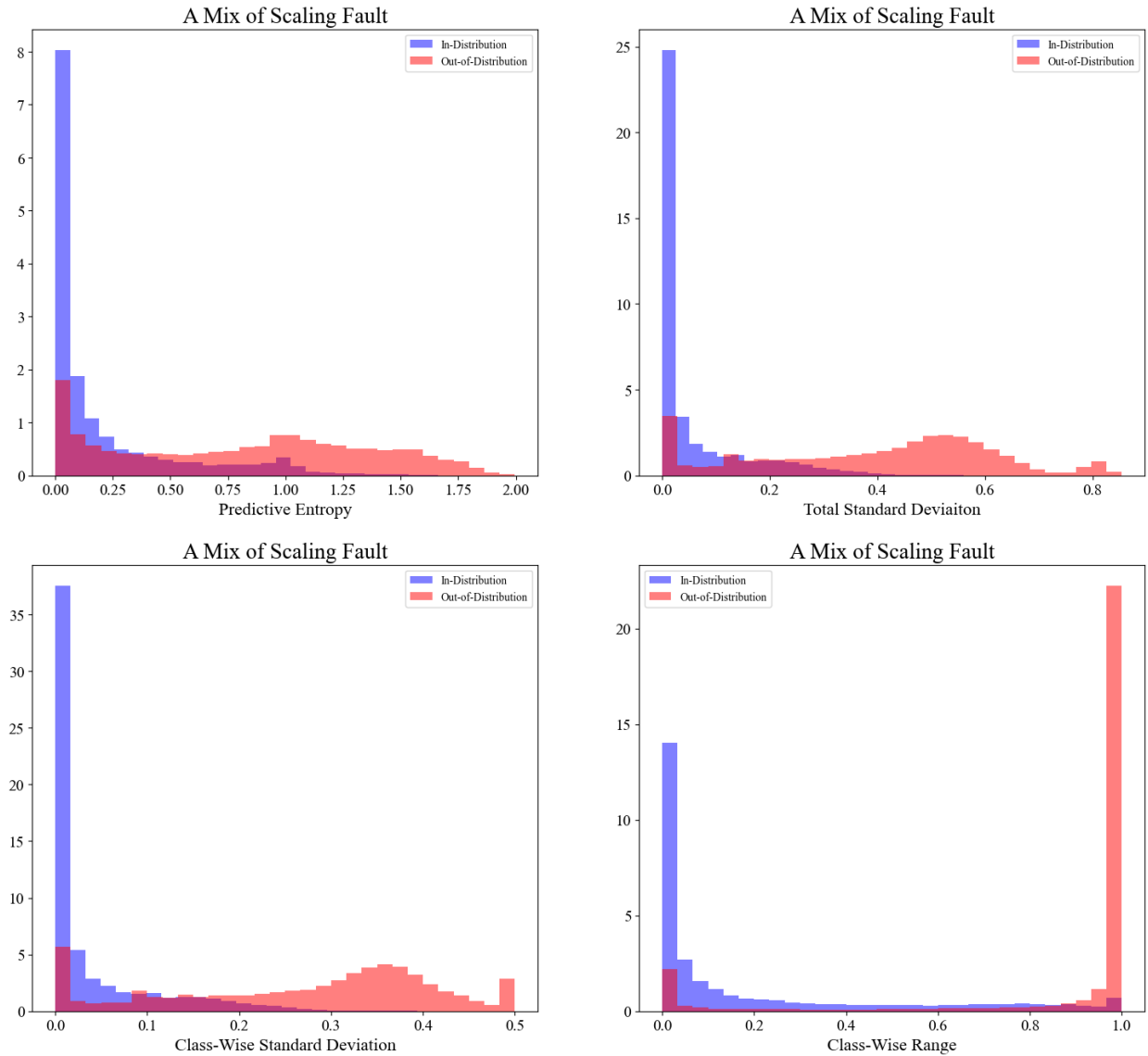

Figure A.5: Comparison of the distributions of predictive uncertainty between in-distribution and out-of-distribution datasets due to scaling fault.

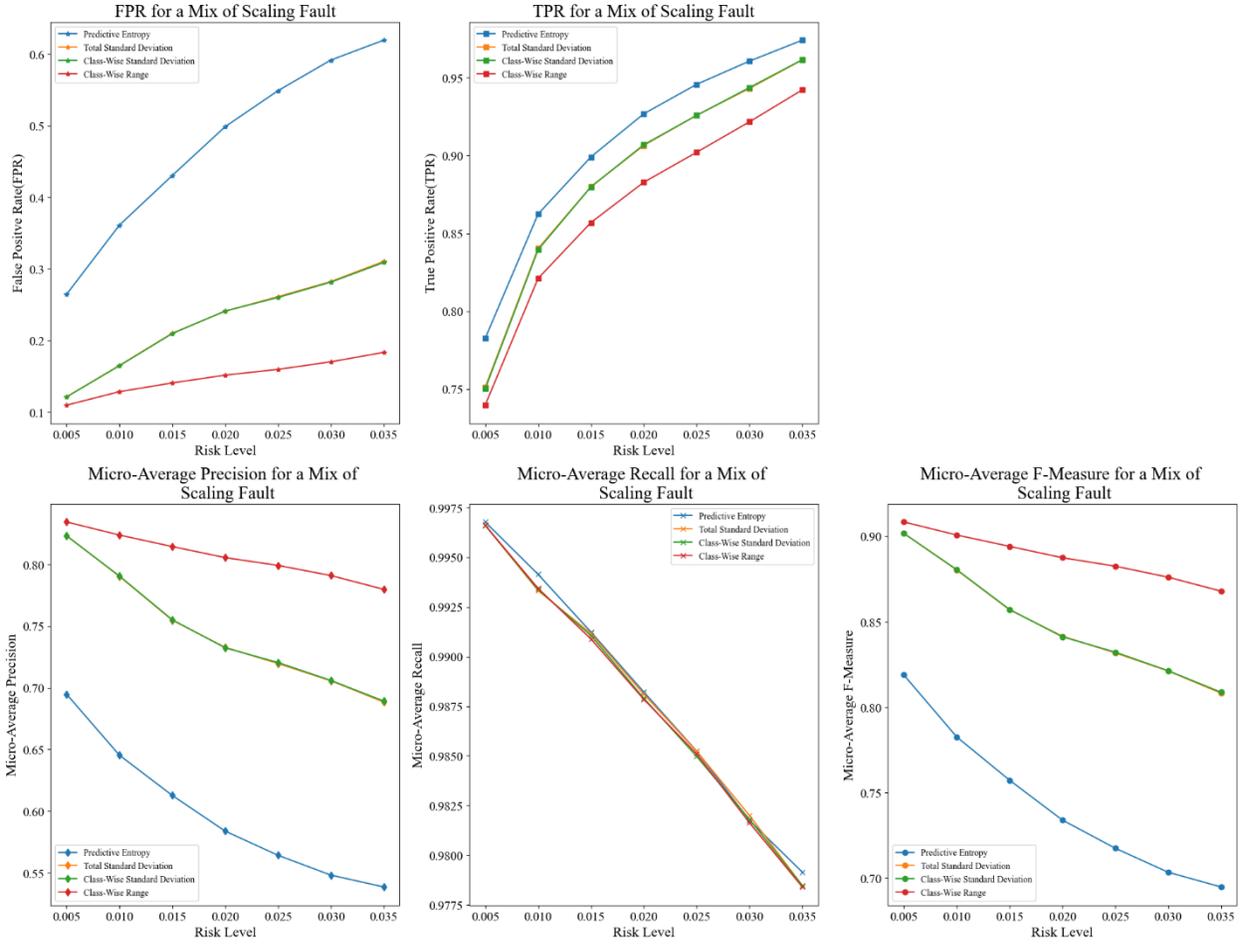

Figure A.6: Model performance of four uncertainty measures in OOD detection and fault diagnosis for a mix of scaling fault under varying risk levels.

## A.4. Precision Degradation.

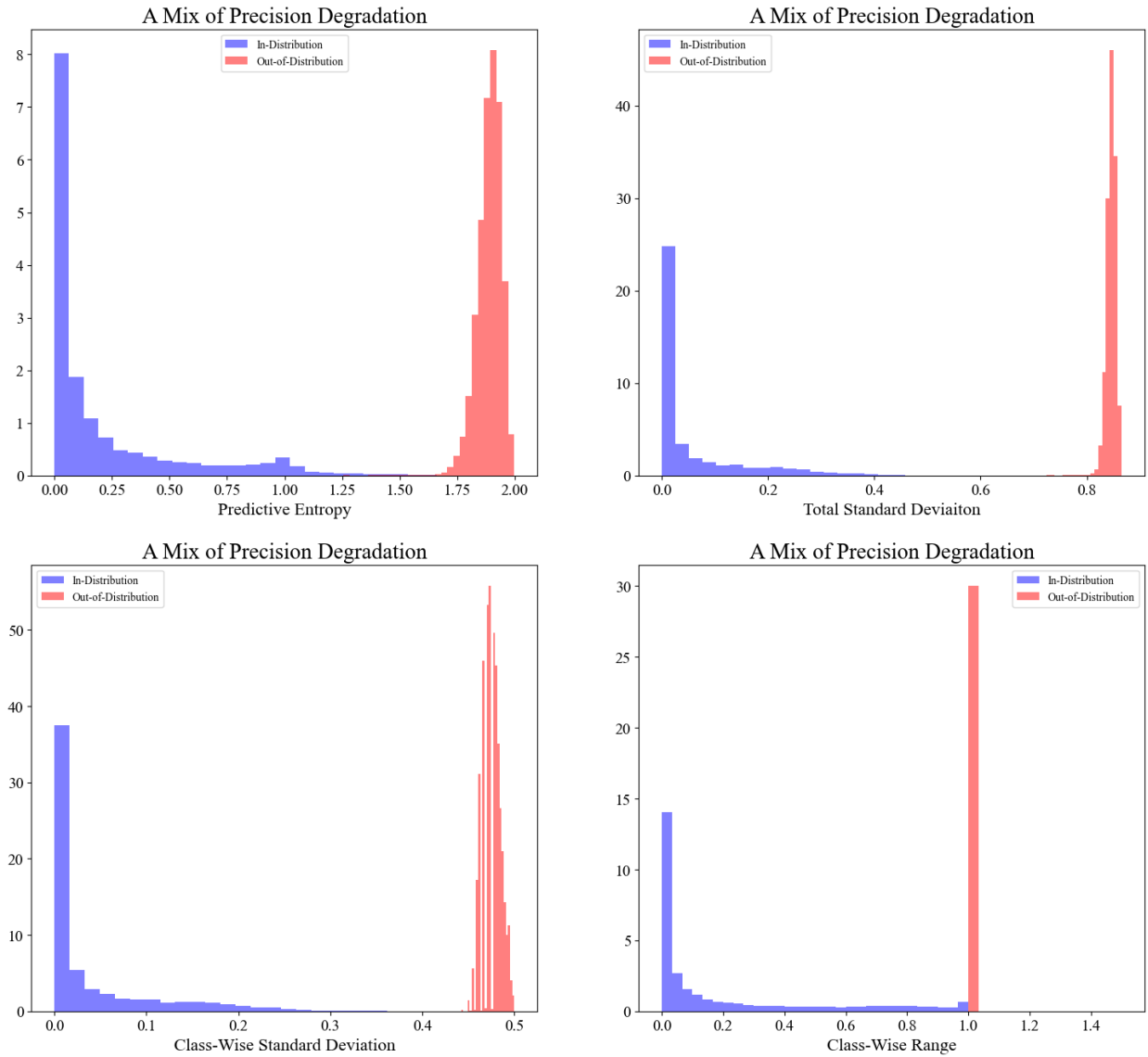

Figure A.7: Comparison of the distributions of predictive uncertainty between in-distribution and out-of-distribution datasets due to precision degradation.

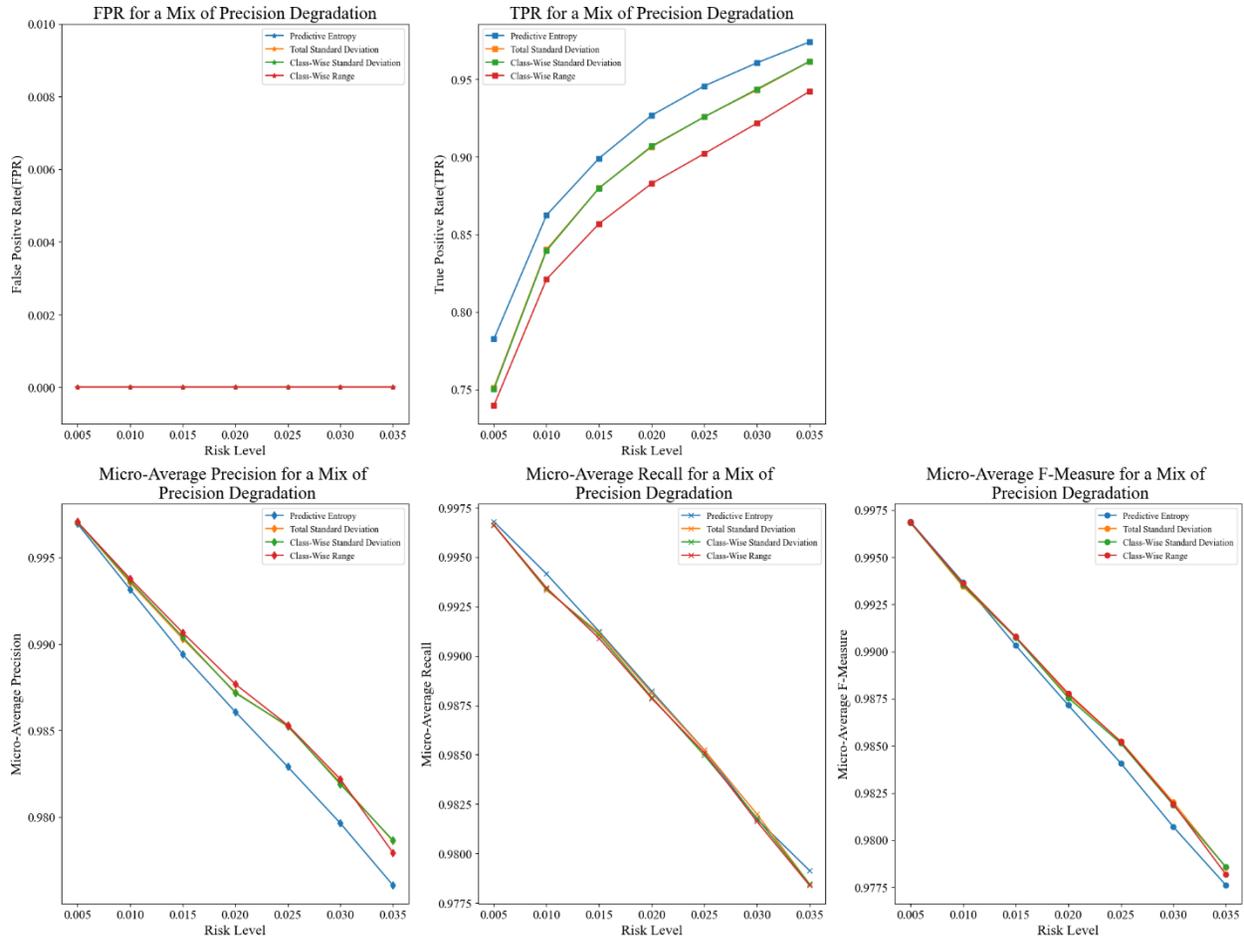

Figure A.8: Model performance of four uncertainty measures in OOD detection and fault diagnosis for a mix of precision degradation under varying risk levels.